\documentclass[12pt]{article} 
\usepackage[utf8]{inputenc}
\usepackage{textgreek}

\usepackage[letterpaper,margin=0.8in]{geometry}

\usepackage{graphicx}
\usepackage{amsmath, amssymb, amsfonts}
\usepackage{amsthm}
\usepackage{mathrsfs}
\usepackage{xcolor}
\usepackage{textcomp}
\usepackage{booktabs}
\usepackage{subcaption}
\usepackage{graphicx}
\usepackage{subcaption}        
\usepackage[percent]{overpic}  
\usepackage[font=small,labelfont=bf]{caption}
\usepackage[scaled=0.92]{helvet}  
\newcommand{\panel}[2]{\textbf{#1}\hspace{0.5em} \includegraphics[width=\linewidth]{#2}}

\usepackage{hyperref}
\usepackage{array}
\usepackage{multirow}
\usepackage{algorithm}
\usepackage{algorithmicx}
\usepackage{algpseudocode}
\usepackage{listings}
\usepackage{array}

\title{Biologically Disentangled Multi-Omic Modeling Reveals Mechanistic Insights into Pan-Cancer Immunotherapy Resistance}
\author{Ifrah Tariq$^{1,2}$ and Ernest Fraenkel$^{1,2,3}$}

\date{\small
$^1$Massachusetts Institute of Technology, Computational and Systems Biology Program, Cambridge, MA, USA \\
$^2$Broad Institute of MIT and Harvard, Cambridge, MA, USA \\
$^3$Massachusetts Institute of Technology, Department of Biological Engineering, Cambridge, MA, USA \\
}

\usepackage{array}
\begin{document}

\maketitle

\begin{abstract}
Immune checkpoint inhibitors (ICIs) have transformed cancer treatment, yet patient responses remain highly variable, and the biological mechanisms underlying resistance are poorly understood. While machine learning models hold promise for predicting responses to ICIs, most existing methods lack interpretability and do not effectively leverage the biological structure inherent to multi-omics data. Here, we introduce the Biologically Disentangled Variational Autoencoder (BDVAE), a deep generative model that integrates transcriptomic and genomic data through modality- and pathway-specific encoders. Unlike existing rigid, pathway-informed models, BDVAE employs a modular encoder architecture combined with variational inference to learn biologically meaningful latent features associated with immune, genomic, and metabolic processes. Applied to a pan-cancer cohort of 366 patients across four cancer types treated with ICIs, BDVAE accurately predicts treatment response (AUC-ROC = 0.94 on unseen test data) and uncovers critical resistance mechanisms, including immune suppression, metabolic shifts, and neuronal signaling. Importantly, BDVAE reveals that resistance spans a continuous biological spectrum rather than strictly binary states, reflecting gradations of tumor dysfunction. Several latent features correlate with survival outcomes and known clinical subtypes, demonstrating BDVAE's capability to generate interpretable, clinically relevant insights. These findings underscore the value of biologically structured machine learning in elucidating complex resistance patterns and guiding precision immunotherapy strategies.
\end{abstract}

\section{Introduction}

Despite the transformative success of immune checkpoint inhibitors (ICIs) in treating a wide range of cancers, patient responses remain highly variable, with many individuals deriving limited or no benefit \cite{darvin2018immune, hossain2024immune}. This variability underscores a critical need to uncover the underlying biological mechanisms of immunotherapy resistance to aid in the development of new drugs and direct patient care to maximize clinical benefit. Machine learning models have shown promise in predicting ICI outcomes; however, most existing approaches fall short in at least one of two key dimensions: interpretability and multi-omics integration \cite{kong2022netbio,vanguri2022multimodal,li2024review}. Many models treat high-dimensional molecular inputs as undifferentiated features, failing to account for known biological structure or cross-modal interactions.

To address this, some efforts have incorporated domain-informed architectures that introduce prior biological knowledge—for example, mapping genes to pathways or introducing concept-level bottlenecks \cite{shen2025compass, elmarakeby2021pnet, greene2021interpretable}. While these designs enhance interpretability by aligning with curated biological knowledge, they often hard-code relationships and limit the model’s ability to generalize across heterogeneous cohorts with varying annotations or batch effects. Moreover, these rigid architectures typically forgo learning a unified latent representation, instead distributing biological abstraction across pre-specified layers. As a result, such models may miss emergent signals and fail to capture the multiscale complexity of tumor-immune dynamics.

To overcome these limitations, we introduce a Biologically Disentangled Variational Autoencoder (BDVAE), a multi-encoder deep generative model that integrates transcriptomic and genomic data while preserving biological structure and interpretability. BDVAE departs from monolithic encoder architectures by assigning separate encoders to biologically coherent feature subsets—such as pathway-level RNA-seq and WES-derived inputs, allowing the model to disentangle distinct axes of variation corresponding to immune, genomic, and metabolic processes. 

Disentanglement, in this context, refers to the ability of the latent space to represent distinct biological factors of variation—such as immune suppression, stromal remodeling, or mutational burden—along independent, interpretable axes. Traditional approaches such as β-VAEs or orthogonality-regularized latent spaces attempt to enforce statistical independence among latent variables. However, these methods are often agnostic to biological semantics and may produce axes that are mathematically decorrelated but biologically uninterpretable. In contrast, BDVAE promotes biological disentanglement by aligning each encoder with pathway-informed feature groupings, which encourages each latent dimension to reflect a distinct biological process, rather than relying solely on statistical constraints. This modular design mitigates the risk of dominant modalities or high-variance features obscuring subtle but biologically meaningful signals, a common pitfall in standard deep learning approaches.

Each encoder learns modality- and pathway-specific embeddings, which are fused into a shared latent space optimized via variational inference. This approach allows the model to capture both known biology—through pathway-specific encoders—and emergent patterns via the unspecified encoder, without imposing hard-coded ontologies. A decoder reconstructs the original inputs, enabling unsupervised learning of latent factors, while a classification head appended to the latent space supports downstream prediction of binary response to immune checkpoint blockade. By aligning the modeling framework with biological priors while maintaining the flexibility of variational inference, BDVAE yields an expressive and interpretable representation that supports mechanistic interpretation, generalization across cohorts, and predictive modeling of immunotherapy response.

\begin{figure}
     \centering
     \includegraphics[width=1\linewidth]{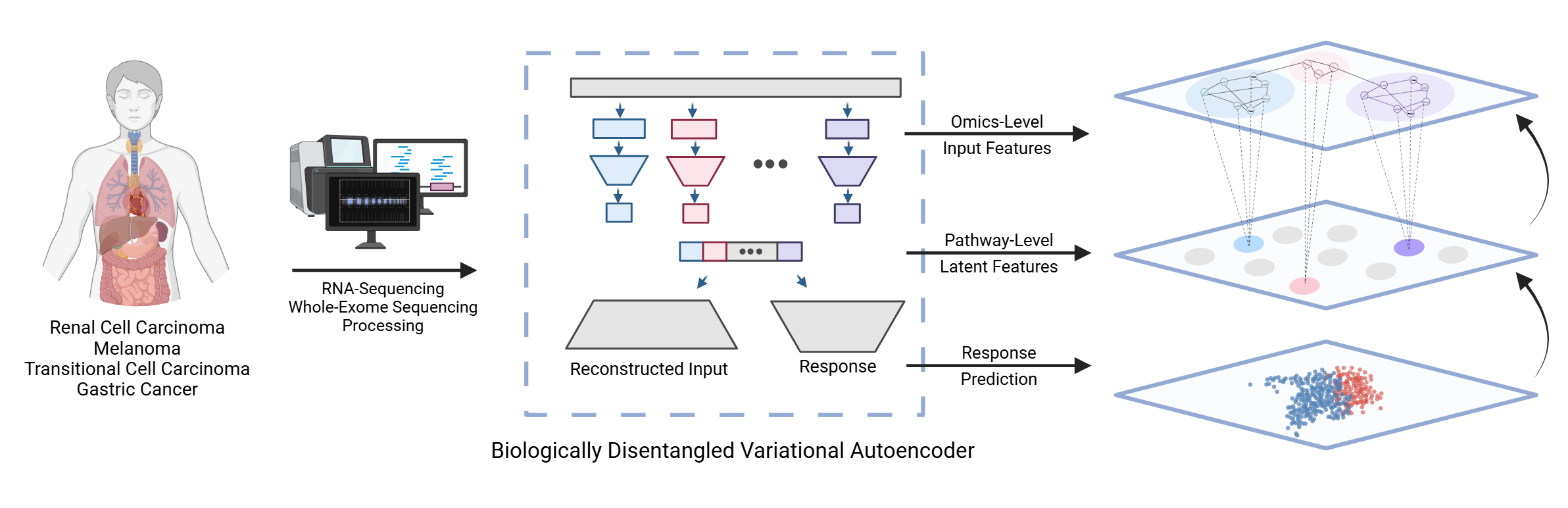}
     \caption{Overview of the Biologically Disentangled Variational Autoencoder (BDVAE) workflow. RNA-sequencing and whole-exome sequencing data from multiple cancer types (renal cell carcinoma, melanoma, transitional cell carcinoma, and gastric cancer) are processed into biologically coherent input features. BDVAE employs separate, pathway-specific encoders to produce interpretable latent representations, capturing distinct biological processes. These latent features facilitate robust reconstruction of input data and enable accurate prediction of patient responses to immune checkpoint blockade therapy. The right panel illustrates how multi-level biological information (omics-level input, pathway-level latent features, and clinical outcomes) is integrated into a unified, interpretable latent space. Created in BioRender. Tariq, I. (2025) https://BioRender.com/svukbrj}
     \label{fig:enter-label}
 \end{figure}

The overview of the BDVAE workflow and architecture is shown in Figure \ref{fig:enter-label}.

\section{Model Notation and Problem Formulation}

Let $\mathbf{x}$ denote the observed data vector for a tumor sample, consisting of concatenated RNA-seq and WES-derived features. The encoder approximates the posterior distribution $q_{\phi}(\mathbf{z}|\mathbf{x})$ over latent variables $\mathbf{z}$, while the decoder reconstructs the input data via a likelihood model $p_{\theta}(\mathbf{x}|\mathbf{z})$. The latent space $\mathbf{z}$ is structured to capture biologically meaningful variation, and the overall model is trained to maximize the evidence lower bound (ELBO) with additional biological regularization terms.

Consider a dataset consisting of $N$ tumor samples, each characterized by genomic and transcriptomic measurements collected before treatment with immune checkpoint inhibitors.  
Each sample belongs to one of $T$ distinct tissue types, and clinical response to immunotherapy is recorded according to RECIST criteria.

Let $\mathbf{x} \in \mathbb{R}^X$ denote the feature vector for a given sample, where $X$ is the total number of input features (genes, mutations, or other biological measurements), and $\mathcal{X} = \{1, \ldots, X\}$ denotes the set of feature indices.  
Each sample is associated with a binary clinical label $y \in \{0,1\}$, indicating treatment response ($y=1$ for responders, $y=0$ for nonresponders).

The objective is to learn a low-dimensional, biologically meaningful latent representation in which different latent variables align with distinct biological processes. In this study, the BDVAE compresses high-dimensional input features ($>10,000$) transcriptomic and genomic measurements into a latent space of $2,000$ dimensions. Unlike traditional statistical disentanglement approaches, the BDVAE framework promotes \textit{biological disentanglement}, encouraging each latent variable to correspond to an independent biological mechanism relevant to immunotherapy response.

We define the overall latent representation as $\mathbf{Z} \in \mathbb{R}^K$, where $K$ denotes the total latent dimensionality.  
The latent vector $\mathbf{Z}$ is constructed by concatenating the outputs of $B$ separate encoders, $\{e_i(\cdot)\}_{i=1}^B$.  
Each encoder $e_i$ produces a $j_i$-dimensional latent vector $\mathbf{z}_i \in \mathbb{R}^{j_i}$:

\begin{equation}
\mathbf{z}_i = e_i(\mathbf{x}_i), \quad i = 1, \ldots, B
\end{equation}

such that:

\begin{equation}
\mathbf{Z} = [\mathbf{z}_1, \mathbf{z}_2, \ldots, \mathbf{z}_B] \in \mathbb{R}^K, \quad \text{where} \quad K = \sum_{i=1}^B j_i.
\end{equation}

Each encoder operates on a masked subset of the full input feature vector.  
Specifically, for each encoder $e_i$, a predefined binary mask $\mathbf{m}_i \in \{0,1\}^X$ selects a subset of features, producing the masked input:

\begin{equation}
\mathbf{x}_i = \mathbf{m}_i \odot \mathbf{x}
\end{equation}

where $\odot$ denotes element-wise multiplication.  
The selected subset of features for encoder $i$ is denoted $\mathcal{X}_i \subset \mathcal{X}$.

For the first $B-U$ encoders, feature masks are defined based on prior biological knowledge, such as tissue-specific gene sets or immune-related pathways.  
The remaining $U$ encoders are not tied to predefined pathways and are instead assigned to capture variation in the residual features. In this study, we used two unspecified encoders—one for transcriptomic features and one for genomic features—ensuring that pathway-unannotated signals from each modality were represented without forcing them into biologically specified groups.  
Specifically, for each encoder $e_i$ with $i = B-U+1, \ldots, B$, the assigned feature set is:

\begin{equation}
\mathcal{X}_i = \mathcal{X} \setminus \bigcup_{j=1}^{B-U} \mathcal{X}_j
\end{equation}

This design enables the model to capture both known biological processes and to discover novel structure among unassigned features.

To reconstruct the original input from the latent space, a shared decoder $D: \mathbb{R}^K \to \mathbb{R}^X$ maps the concatenated latent vector $\mathbf{Z}$ back to the full input feature space:

\begin{equation}
\hat{\mathbf{x}} = D(\mathbf{Z})
\end{equation}

where $\hat{\mathbf{x}} \in \mathbb{R}^X$ represents the reconstructed input.

In parallel, a classifier $C: \mathbb{R}^K \to [0,1]$ operates on the latent space to predict the probability of treatment response.  
This dual objective ensures that the learned latent variables are both representative of the original biological data and predictive of clinical outcomes.

\subsection{Loss Function}

The BDVAE is optimized by minimizing a composite loss function that balances three competing objectives: accurate reconstruction of the input data, regularization of the latent space, and prediction of clinical treatment response.  
Formally, the total loss $\mathcal{L}_0$ is defined as:

\begin{equation}
\mathcal{L}_0 
= \lambda_{\text{rec}} \, \mathcal{L}_{\text{rec}} 
+ \lambda_{\text{MMD}} \, \mathcal{L}_{\text{MMD}} 
+ \lambda_{\text{resp}} \, \mathcal{L}_{\text{resp}}
\end{equation}

where:
\begin{itemize}
    \item $\lambda_{\text{rec}} \, \mathcal{L}_{\text{rec}}$ denotes the weighted reconstruction loss, encouraging accurate recovery of the original input features. 
    \item $\lambda_{\text{MMD}} \, \mathcal{L}_{\text{MMD}}$ denotes the weighted Maximum Mean Discrepancy loss, promoting desirable distributional properties in the latent space. 
    \item $\lambda_{\text{resp}} \, \mathcal{L}_{\text{resp}}$ denotes the weighted binary cross-entropy loss for predicting immunotherapy treatment response.
\end{itemize}

Reconstruction loss is computed using mean squared error (MSE) for continuous RNA expression features and binary cross-entropy (BCE) for discrete mutation presence features.  
Each loss component was weighted equally during training unless otherwise specified.  
Sensitivity analyses confirmed that moderate adjustments to these weights did not qualitatively alter the learned latent structure or downstream biological interpretations.

\subsection{Model Outputs and Interpretation}

Upon completion of training, the BDVAE produces several outputs for each sample that support downstream biological and clinical analyses.

First, the model generates latent embeddings, compressing the input features by at least 5-fold, providing lower-dimensional representations that disentangle distinct biological processes associated with immunotherapy response. 

Second, it produces reconstructed features, including predicted RNA expression values and mutation presence indicators, which are used to assess reconstruction fidelity.  
Third, the BDVAE yields latent distribution parameters, namely the mean and variance estimates for each latent dimension, enabling downstream analyses such as uncertainty quantification and sample clustering.  

Finally, the model supports feature attribution by enabling post hoc interpretation of the learned latent features using SHAP and Integrated Gradients methods, linking input features to latent biological factors.

Together, this design aligns latent dimensions with known biological programs while preserving flexibility for novel pathway discovery. BDVAE is benchmarked against standard baselines, tested across cancer types, and interrogated at three levels of interpretation: response-level patient stratification, pathway-level signaling programs, and gene-level therapeutic targets.

\section{Results}

\subsection{Study Cohort and Clinical Response Annotation}

Following RNA-seq and WES preprocessing, clinical harmonization, and quality control, a total of 366 tumor samples across four cancer types were retained for analysis. Each sample was labeled as a responder or non-responder based on clinical outcome following immune checkpoint inhibitor (ICI) therapy.

The distribution of samples by cancer type and response category is summarized in Table~\ref{tab:sample_summary}. 

\begin{table}[ht]
\centering
\caption{Sample distribution by cancer type and immunotherapy response status. The full cohort ($N=366$) was split into training ($N=234$), validation ($N=59$), and test ($N=73$) sets using stratified sampling to preserve cancer type and response proportions.}
\begin{tabular}{|l|c|c|c|}
\hline
\textbf{Cancer Type} & \textbf{Responders (N)} & \textbf{Non-Responders (N)} & \textbf{Total (N)} \\
\hline
Gastric & 8 & 28 & 36 \\
Melanoma & 28 & 77 & 105 \\
Renal Cell Carcinoma (RCC) & 32 & 85 & 118 \\
Transitional Cell Carcinoma (TCC) & 27 & 80 & 107 \\
\hline
\textbf{Total} & \textbf{95} & \textbf{270} & \textbf{366} \\
\hline
\end{tabular}

\vspace{0.5em}
\label{tab:sample_summary}
\end{table}

\subsection{BDVAE Latent Representations Enable Accurate and Interpretable Prediction}

\begin{figure}[H]
  \centering
  \begin{subfigure}[t]{0.485\textwidth}
  \panel{A}{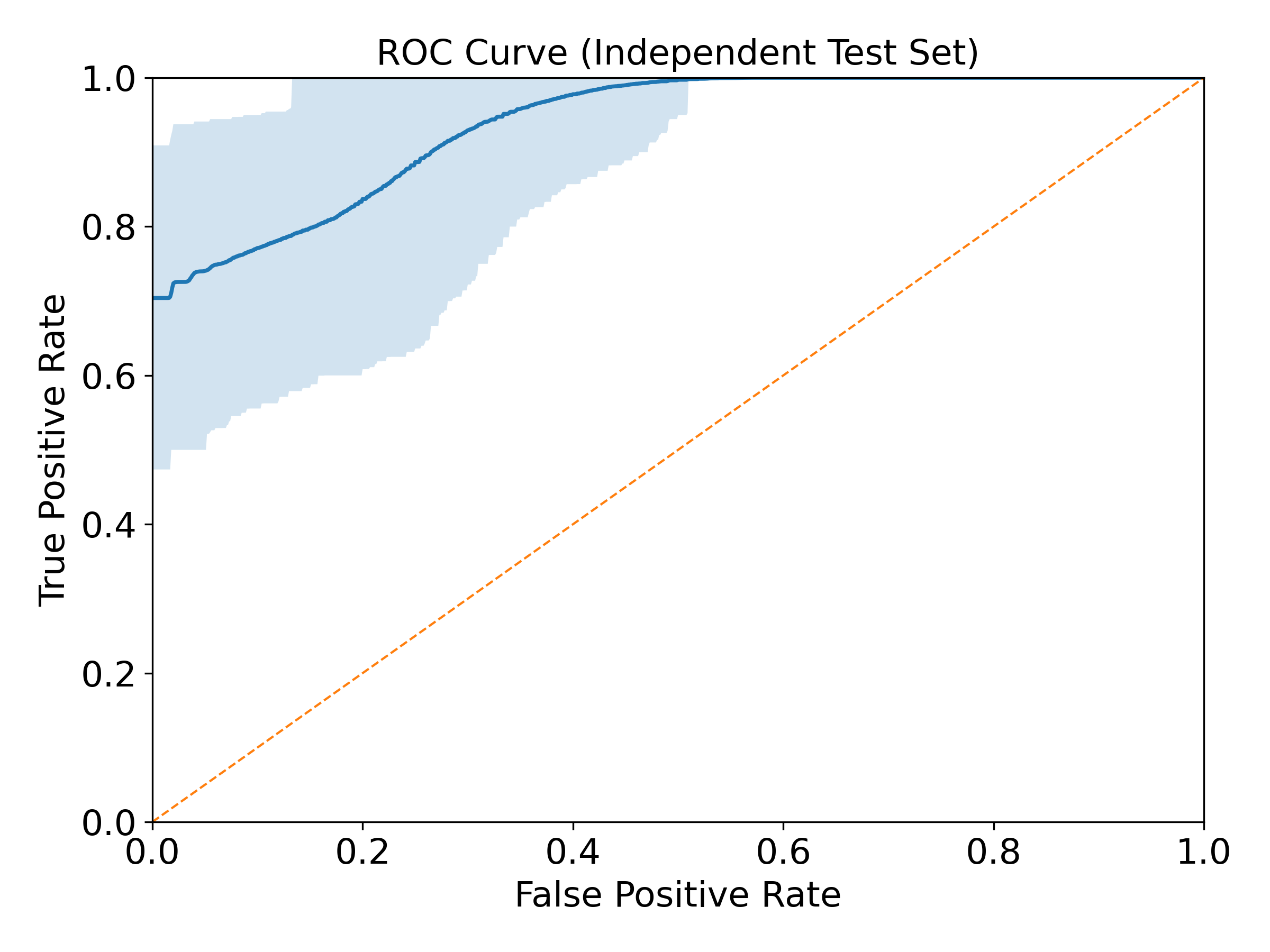}
  \end{subfigure}\hspace{0.02\textwidth}
  \begin{subfigure}[t]{0.485\textwidth}
  \panel{B}{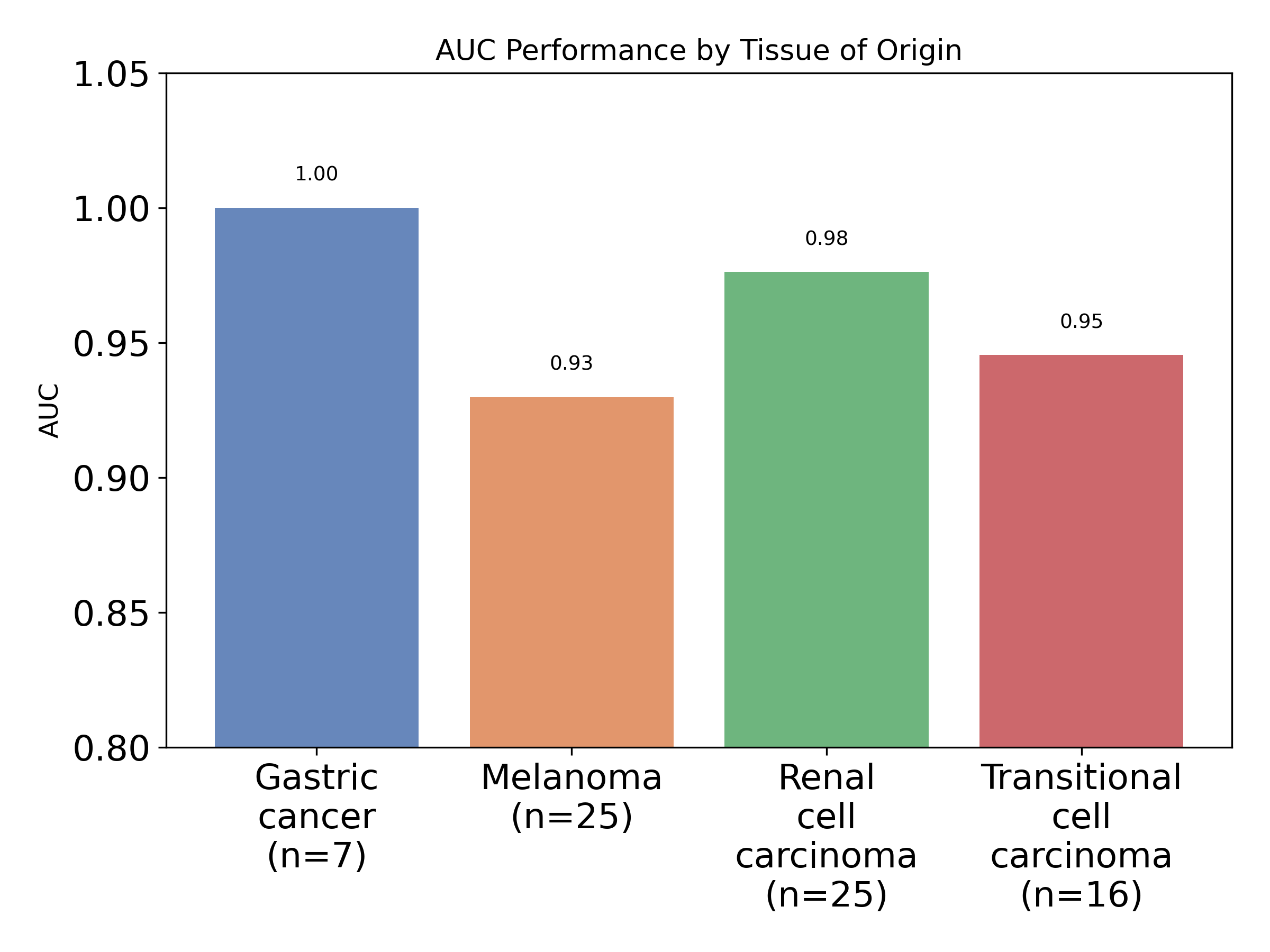}
  \end{subfigure}

  \vspace{0.6em}

  \begin{subfigure}[t]{0.99\textwidth}
  \panel{C}{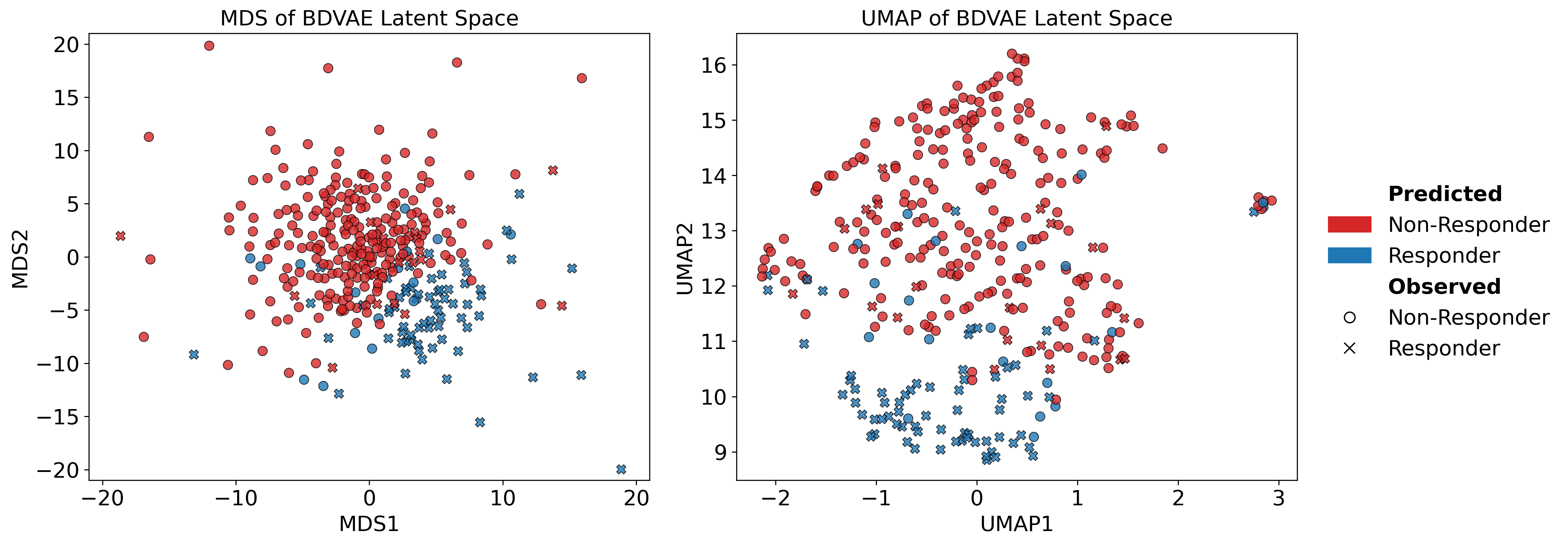}
  \end{subfigure}

  \caption{(A) Receiver operating characteristic (ROC) curve on an independent test set, with 95\% confidence interval shown in blue shading.
    (B) Area under the ROC curve (AUC) stratified by tissue of origin, showing consistently high predictive performance across gastric cancer, melanoma, renal cell carcinoma, and transitional cell carcinoma cohorts in the test set.
    (C) Multidimensional scaling (MDS) projection of BDVAE latent representations (left panel), colored by responder and non-responder labels, revealing separation between outcome groups. Uniform Manifold Approximation and Projection (UMAP) embedding of BDVAE latent representations (right panel), colored by predicted and true labels, demonstrating clustering of responders and non-responders in the learned latent space for all samples.}
     \label{fig:latent_space}
\end{figure}

BDVAE-derived latent features were used to train a LightGBM classifier, which leverages gradient-boosted decision trees to capture non-linear feature interactions in the latent space. We evaluated the predictive performance of the BDVAE–LightGBM pipeline on an independent test set comprising four cancer types. The model demonstrated robust discrimination of responders from non-responders, achieving an overall area under the ROC curve (AUC-ROC) of 0.94 (95\% CI: 0.87–0.99) (Figure~\ref{fig:latent_space}A). Bootstrap resampling confirmed the stability of the ROC estimate, with narrow confidence intervals across most of the false-positive rate range. These results highlight the strong generalization capacity of BDVAE-derived representations when applied to unseen patients.

To further evaluate model generalizability across tumor types, we examined ROC performance stratified by tissue of origin (Figure~\ref{fig:latent_space}B). High predictive accuracy was observed across all four cohorts: gastric cancer (AUC = 1.00, $n=7$), melanoma (AUC = 0.93, $n=25$), renal cell carcinoma (AUC = 0.98, $n=25$), and transitional cell carcinoma (AUC = 0.95, $n=16$). These consistently high AUCs indicate that the model captures shared biological signals underlying immunotherapy response while maintaining predictive robustness across diverse tumor contexts. Notably, even in the small gastric cancer cohort, the model achieved perfect separation of responders and non-responders. Nevertheless, evaluation on additional independent cohorts will be required to further confirm the model’s generalizability.

Having established strong predictive performance across test cohorts, we next asked whether BDVAE latent features encode clinically meaningful variation independent of supervised training. To this end, we evaluated the separation between responders and non-responders in latent space using two complementary distributional metrics: Energy Distance and Maximum Mean Discrepancy (MMD). Both Energy Distance and MMD tests revealed highly significant differences between responder and non-responder embeddings (Energy Distance = 2.072, $p < 0.0001$; MMD = 0.0160, $p < 0.0001$). Null distributions and observed statistics are provided in Supplementary Figure~\ref{fig:supp_latent_permutation}.

We further visualized the latent embeddings using multidimensional scaling (MDS) and uniform manifold approximation and projection (UMAP). In both projections, partial but discernible clustering of samples by response status was observed (Figure~\ref{fig:latent_space}C). Importantly, misclassified samples tended to localize near the boundary between responder- and non-responder–enriched regions, consistent with ROC analyses. These findings confirm that BDVAE not only enables accurate prediction but also captures clinically meaningful structure in latent space.

\subsection{Biological Interpretation of BDVAE Latent Representation}

Having established that BDVAE-derived features enable robust and generalizable prediction of immunotherapy response across multiple tumor types, we next sought to interpret the latent representations to gain mechanistic insight. Specifically, we asked whether the model’s learned latent dimensions align with clinically and biologically meaningful programs. To address this, we ranked latent features by their contribution to response prediction, examined their enrichment for established or novel pathways, and assessed latent space organization through clustering and projection analyses.

\begin{figure}[ht]
    \centering
    \includegraphics[width=0.8\textwidth]{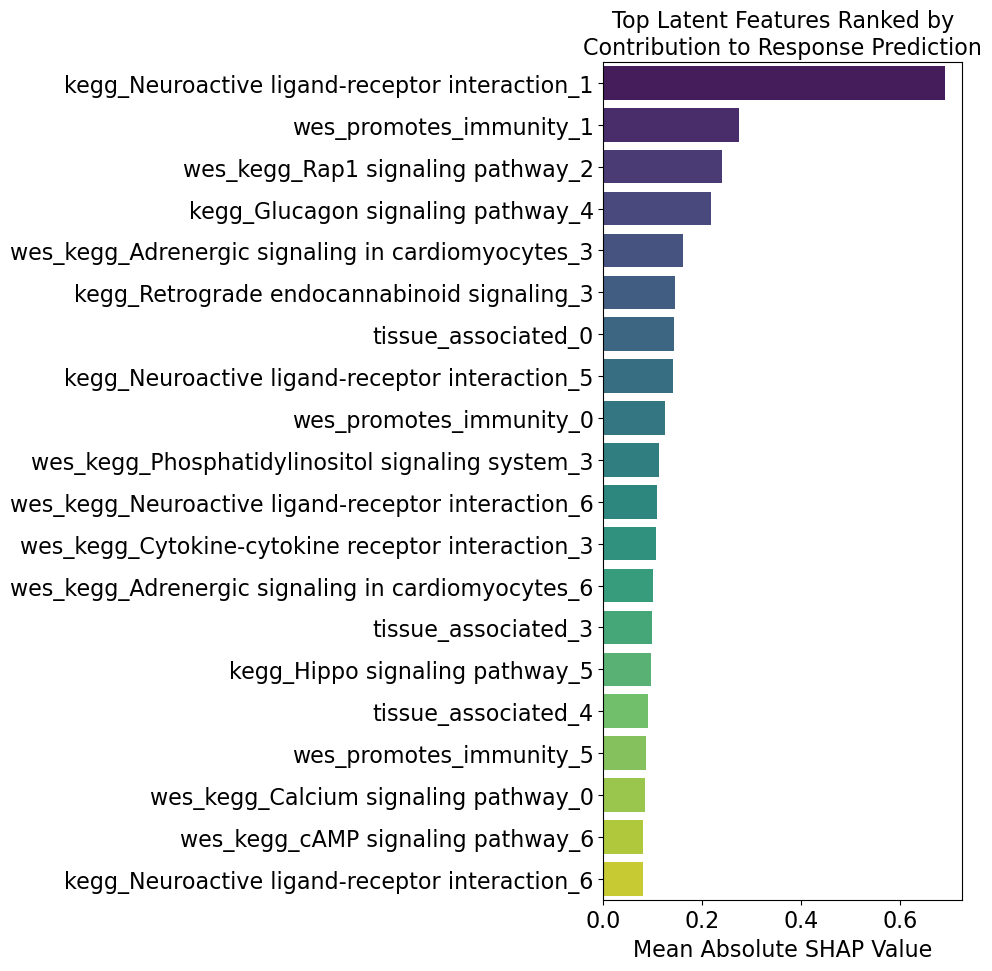}
    \caption{
    Top-ranked latent dimensions by mean absolute SHAP value. Latent names beginning with ``wes'' indicate WES-derived features, while others are RNA-derived. 
  Immune-related programs and tumor-intrinsic pathways dominate model predictions of immunotherapy response.}
    \label{fig:top_latents_shap}
\end{figure}

To interpret which features most strongly influenced response prediction, we applied SHAP analysis to the BDVAE–LightGBM classifier. As shown in Figure~\ref{fig:top_latents_shap}, the most influential latent dimensions were enriched for pathways involved in immune regulation, tumor-intrinsic signaling, and neuroimmune interactions. The top-ranked feature corresponded to the \textit{Neuroactive Ligand–Receptor Interaction} pathway, followed by latent dimensions linked to immune-promoting cytokine programs, Rap1 signaling, and adrenergic pathways. These results support the view that BDVAE not only achieves strong predictive performance but also disentangles biologically interpretable axes underlying clinical outcomes.

\begin{figure}[!htbp]
    \raggedright   
  \begin{subfigure}[t]{0.7\textwidth}
  \panel{A}{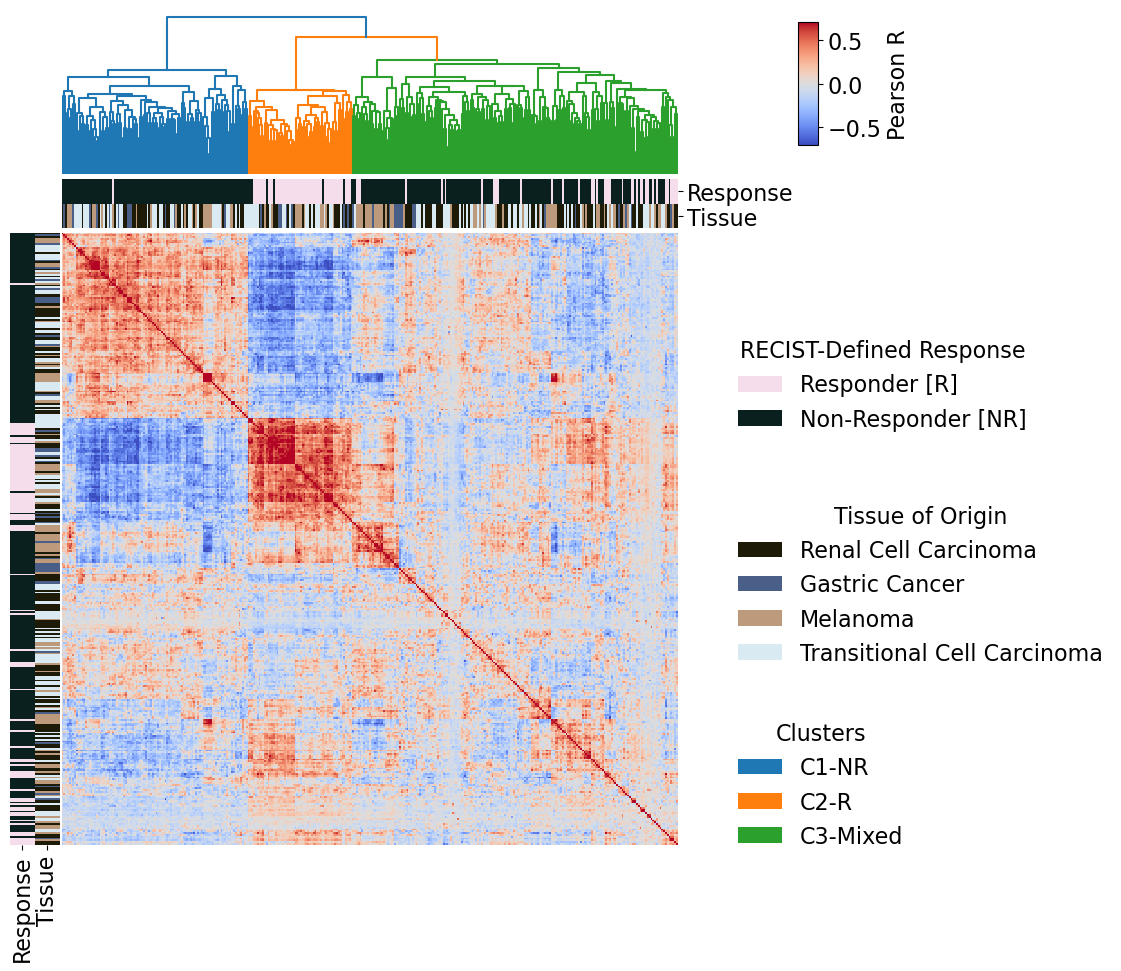}
  \end{subfigure}\hspace{0.02\textwidth}
  \begin{subfigure}[t]{0.485\textwidth}
  \panel{B}{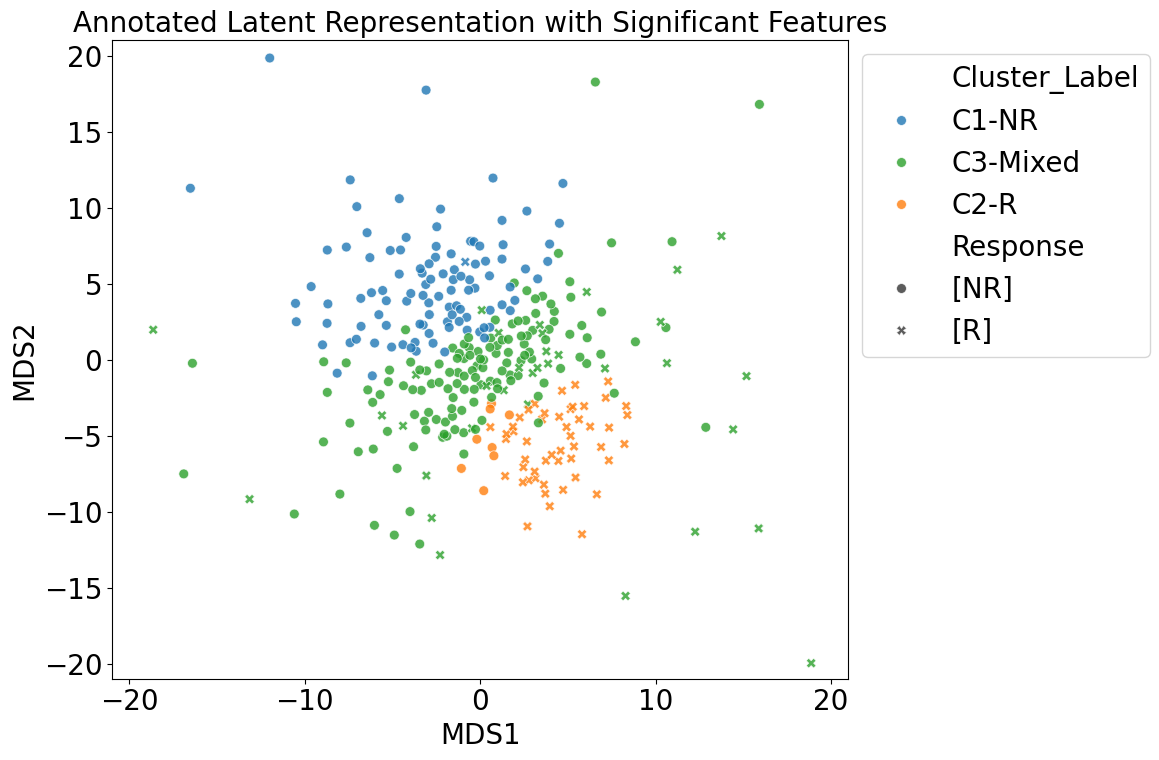}
  \end{subfigure}\hspace{0.02\textwidth}
  \begin{subfigure}[t]{0.485\textwidth}
  \panel{C}{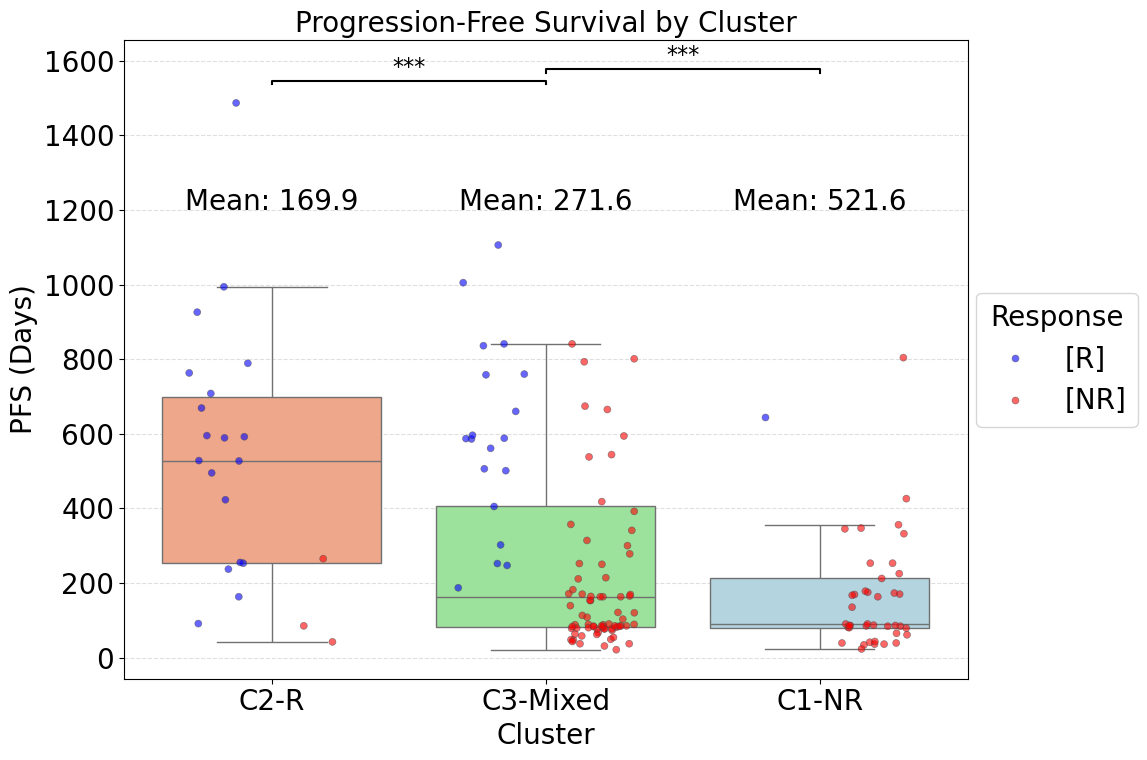}
  \end{subfigure}

\caption{
\textbf{BDVAE latent structure reveals distinct clinical subgroups.} 
(A) Hierarchical clustering of pairwise correlations in the BDVAE latent space identified three subgroups: C1-NR (non-responder–enriched), C2-R (responder-enriched), and C3-Mixed. 
(B) MDS projection annotated by cluster membership, showing clear alignment between latent clusters and response status.
(C) Progression-free survival (PFS) stratified by cluster membership, with C2-R patients exhibiting significantly longer survival. }
\label{fig:latent_clusters_pfs}
\end{figure}

We next examined whether these significant features organized patients into clinically meaningful subgroups. We first standardized all significant latent features to zero mean and unit variance. A sample--sample Pearson correlation matrix was then computed, capturing pairwise similarities among tumors in the refined latent space. Hierarchical clustering with average linkage was applied to the correlation matrix, and clusters were defined by cutting the dendrogram at a fixed distance threshold. Hierarchical clustering of pairwise correlations in the refined latent space revealed three major subgroups: C1-NR (non-responder dominant), C2-R (responder dominant), and C3-Mixed (heterogeneous with mixed response profiles). Importantly, these latent clusters were clinically relevant: patients in C2-R experienced significantly longer progression-free survival compared to C1-NR and C3-Mixed (Figure~\ref{fig:latent_clusters_pfs}B). Patients in the C3-Mixed cluster had an intermediate PFS compared to either C1-NR and C2-R. Multidimensional scaling projections further confirmed that the three clusters occupy distinct regions of the latent embedding (Figures~\ref{fig:latent_clusters_pfs}C–D), with C1-NR and C2-R aligning closely with non-responder and responder–enriched regions, respectively. Together, these findings demonstrate that the BDVAE latent space organizes tumors along a biologically interpretable axis of immunotherapy response, with an intermediate group reflecting mixed or transitional molecular phenotypes.

\subsubsection{Pathway-Level Interpretation of Latent Features}

\begin{figure}[htbp ]
    \centering
    \includegraphics[width=0.95\textwidth]{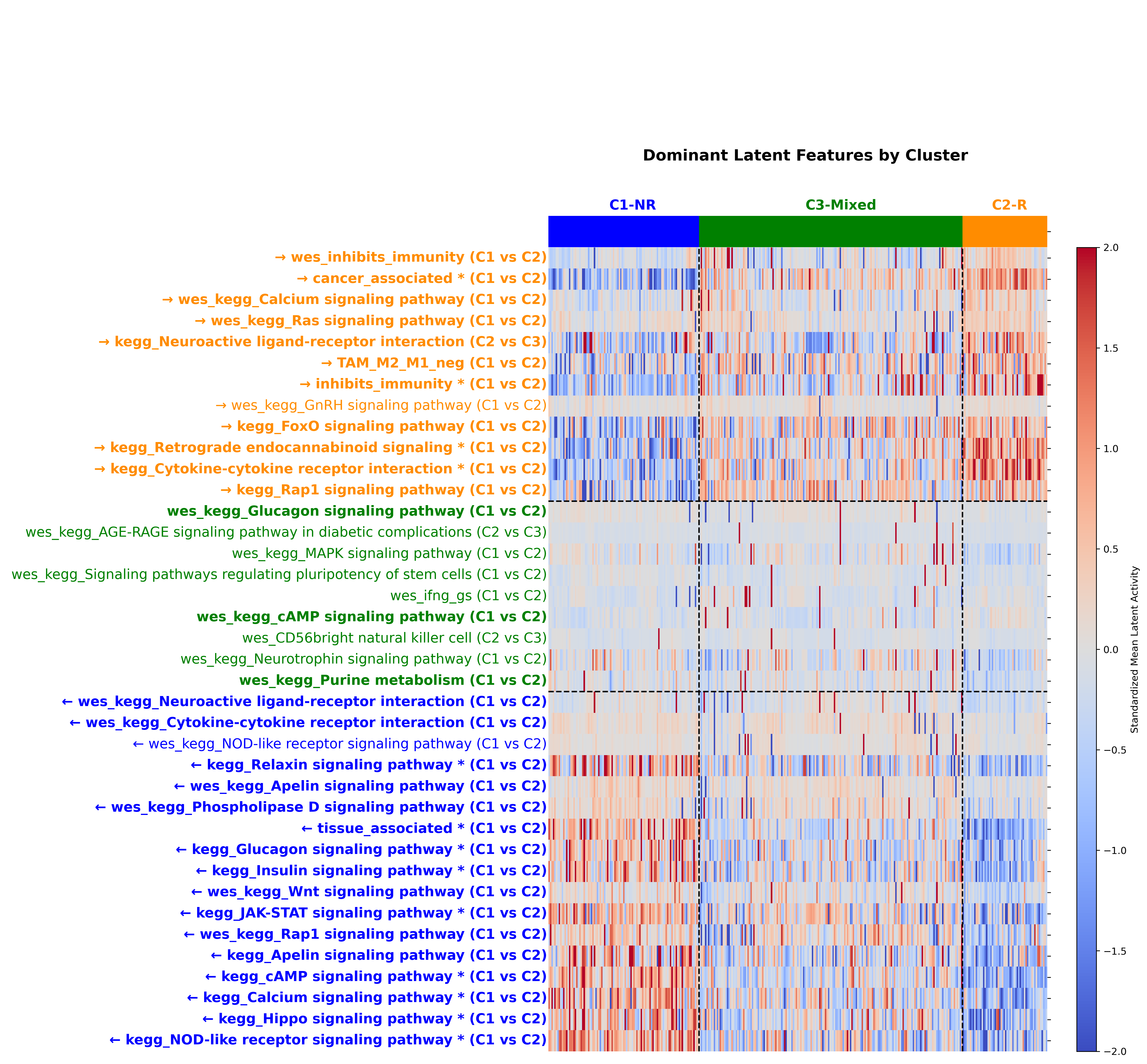}
    \caption{
    \textbf{Pathway-level interpretation of BDVAE latent features.} 
    Heatmap of standardized pathway activity across tumor samples, grouped by latent cluster (C1-NR, C3-Mixed, C2-R). 
    Rows correspond to pathways, annotated by their dominant enrichment direction ($\rightarrow$ C2-R, $\leftarrow$ C1-NR, $\leftrightarrow$ C3-Mixed). 
    Features with large effect sizes ($|\delta| > 0.47$) are bolded; very strong effects ($|\delta| > 0.7$) are marked with an asterisk. 
    }
    \label{fig:latent_pathway_heatmap}
\end{figure}

To uncover biological programs encoded by BDVAE, we grouped latent dimensions by their respective encoder annotations (e.g., \texttt{TAM\_M2\_m1\_neg}, \texttt{kegg\_Insulin\_Signaling\_pathway}). Within each patient cluster (C1-NR, C3-Mixed, C2-R), pathway activity was summarized as the median standardized activation across associated latents. Pathways with a maximum pairwise cluster difference exceeding 0.03 were retained. To quantify separation, we computed Cliff’s delta between clusters, bolding features with large effect sizes ($|\delta| > 0.47$) and marking those with very strong separation ($|\delta| > 0.7$) with an asterisk (Figure~\ref{fig:latent_pathway_heatmap}).

\subsubsection{Biological Programs Underlying Cluster-Dominant Latent Features}

To further refine biological interpretation, we categorized dominant latent features into four broad functional groups: 
\begin{enumerate}
    \item immune-related programs,
    \item cancer-associated signaling,
    \item neuronal pathways, and
    \item metabolic or miscellaneous processes.
\end{enumerate}
Each cluster exhibited a distinct combination of these biological programs, reflecting divergent mechanisms of immunotherapy response and resistance.

\paragraph{C1-NR (Non-Responder Dominant Cluster)}  
\textbf{Dominant Latent Features:}
\begin{itemize}
  \item \textbf{Immune:} Cytokine-cytokine receptor interaction, NOD-like receptor signaling, JAK-STAT signaling
  \item \textbf{Cancer-associated:} Relaxin signaling, Apelin signaling, Wnt signaling, Rap1 signaling, Hippo signaling
  \item \textbf{Neuronal:} Neuroactive ligand-receptor interaction, Phospholipase D signaling, cAMP signaling, Calcium signaling
  \item \textbf{Metabolic/Miscellaneous:} Glucagon signaling, Insulin signaling, Tissue-specific pathways
\end{itemize}

\paragraph{C2-R (Responder Dominant Cluster)}  
\textbf{Dominant Latent Features:}
\begin{itemize}
  \item \textbf{Immune:} Inhibitory immune signatures, TAM M2/M1 polarization signatures, Cytokine-cytokine receptor interaction
  \item \textbf{Cancer-associated:} Ras signaling, FoxO signaling, Rap1 signaling, General cancer pathways
  \item \textbf{Neuronal:} Calcium signaling, Neuroactive ligand-receptor interaction, GnRH signaling, Retrograde endocannabinoid signaling
\end{itemize}

\paragraph{C3-Mixed (Intermediate Cluster)}  
\textbf{Dominant Latent Features:}
\begin{itemize}
  \item \textbf{Immune:} IFNG-related signatures, CD56\textsuperscript{bright} NK cell activation
  \item \textbf{Cancer-associated:} MAPK signaling, Signaling pathways regulating pluripotency of stem cells
  \item \textbf{Neuronal:} cAMP signaling, Neurotrophin signaling, Purine metabolism
  \item \textbf{Metabolic/Miscellaneous:} Glucagon signaling, AGE-RAGE signaling (diabetic complications)
\end{itemize}

These results suggest that BDVAE disentangles clinically relevant biological programs across immune activation, tumor-intrinsic signaling, neuronal modulation, and metabolic adaptation. Together, the three clusters capture both discrete clinical phenotypes and transitional molecular states.  

This overview highlights the dynamic interplay of immune, tumor-intrinsic, and neuronal pathways characterizing each subgroup. To develop a deeper appreciation of the molecular distinctions associated with clinical outcomes, we next focus on a direct comparison between the responder-dominant (C2-R) and non-responder-dominant (C1-NR) clusters, followed by a closer examination of the intermediate features in Cluster~3 (C3-Mixed).

\subsection{Biological Pathways Underlying Clinical Response and Resistance }

To isolate molecular programs most directly associated with clinical outcome, we compared latent feature distributions between C1-NR (non-responder dominant) and C2-R (responder dominant), excluding the heterogeneous C3-Mixed group. Latent features were ranked by absolute median differences between clusters, and Cliff’s delta was computed to prioritize distinguishing signals. Features were grouped by directionality (upregulated in responders or non-responders) and visualized as standardized heatmaps. 

This analysis revealed sharp molecular contrasts: responders were enriched for immune-activating and modulatory pathways, whereas non-responders exhibited heightened inflammatory, neuroimmune, and developmental signaling programs.

\subsubsection{Immune Signaling Profiles Differentiating Responders and Non-Responders}

To identify immune pathways driving separation between responder and non-responder phenotypes, we computed gene-level attribution scores using integrated gradients and compared their distributions across clusters. Supplementary Figure~\ref{fig:immune_volcano_panels} displays volcano plots for eight major immune-related pathways, including \textit{Cytokine–Cytokine Receptor Interaction}, \textit{NOD-like Receptor Signaling}, \textit{JAK–STAT Signaling}, and \textit{TAM M2/M1 negative signature}.

\begin{figure}[H]
    \centering
    \includegraphics[width=0.95\textwidth]{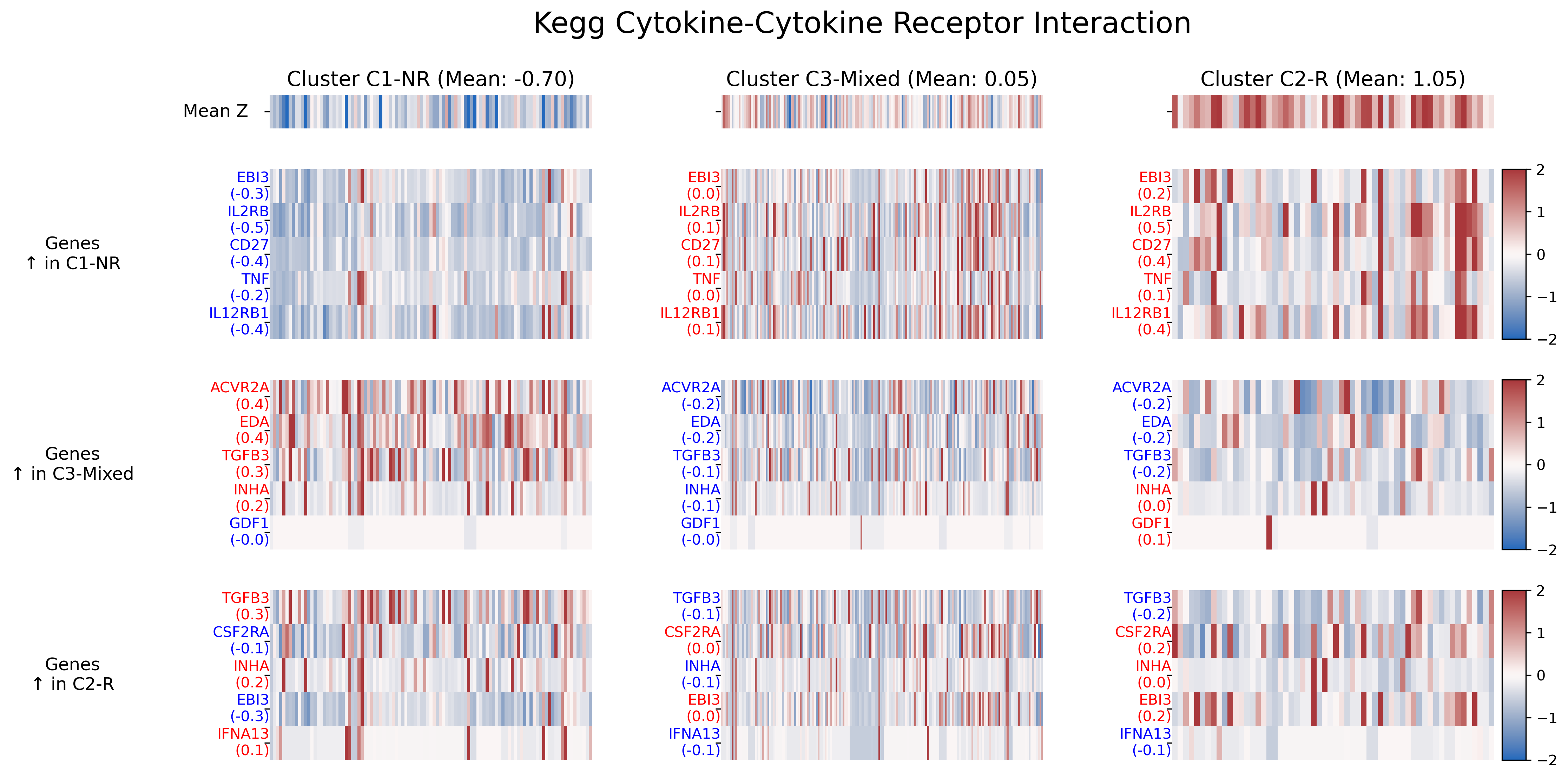}
    \caption{
    Heatmap of z-scored gene expression across clusters for selected genes in the \textit{Cytokine-Cytokine Receptor Interaction} pathway.
    Columns correspond to tumor samples grouped by cluster; rows represent key genes annotated by their expression direction and mean z-score.
    Genes such as \textit{CXCL9} and \textit{CXCR3} show higher expression in C2-R, while \textit{TGFB3} and \textit{IFNA13} are more active in C1-NR.
    Expression in Cluster~3 (C3-Mixed) is intermediate across several genes, suggesting a gradient or transitional immune phenotype.
    }
    \label{fig:cytokine_expression_heatmap}
\end{figure}

To investigate immunological pathways associated with differential therapeutic outcomes, we examined attribution and expression patterns for genes in the \textit{Cytokine–Cytokine Receptor Interaction} pathway. We analyzed gene-level z-scores for the top contributing cytokine pathway genes across clusters. As shown in Figure~\ref{fig:cytokine_expression_heatmap}, responder tumors exhibited elevated expression of \textit{CXCR3}, \textit{CXCL9}, and \textit{CD40LG}, while non-responders showed consistently higher expression of \textit{TGFB3}, \textit{IL1R2}, and \textit{EBI3}. Notably, many genes—including \textit{CXCR3}, \textit{CXCL9}, and \textit{IL18}—exhibited a smooth gradient across clusters, with C3-Mixed tumors showing intermediate expression levels, further supporting a transitional immune state.

\begin{figure}[H]
    \centering
    \includegraphics[width=0.6\textwidth]{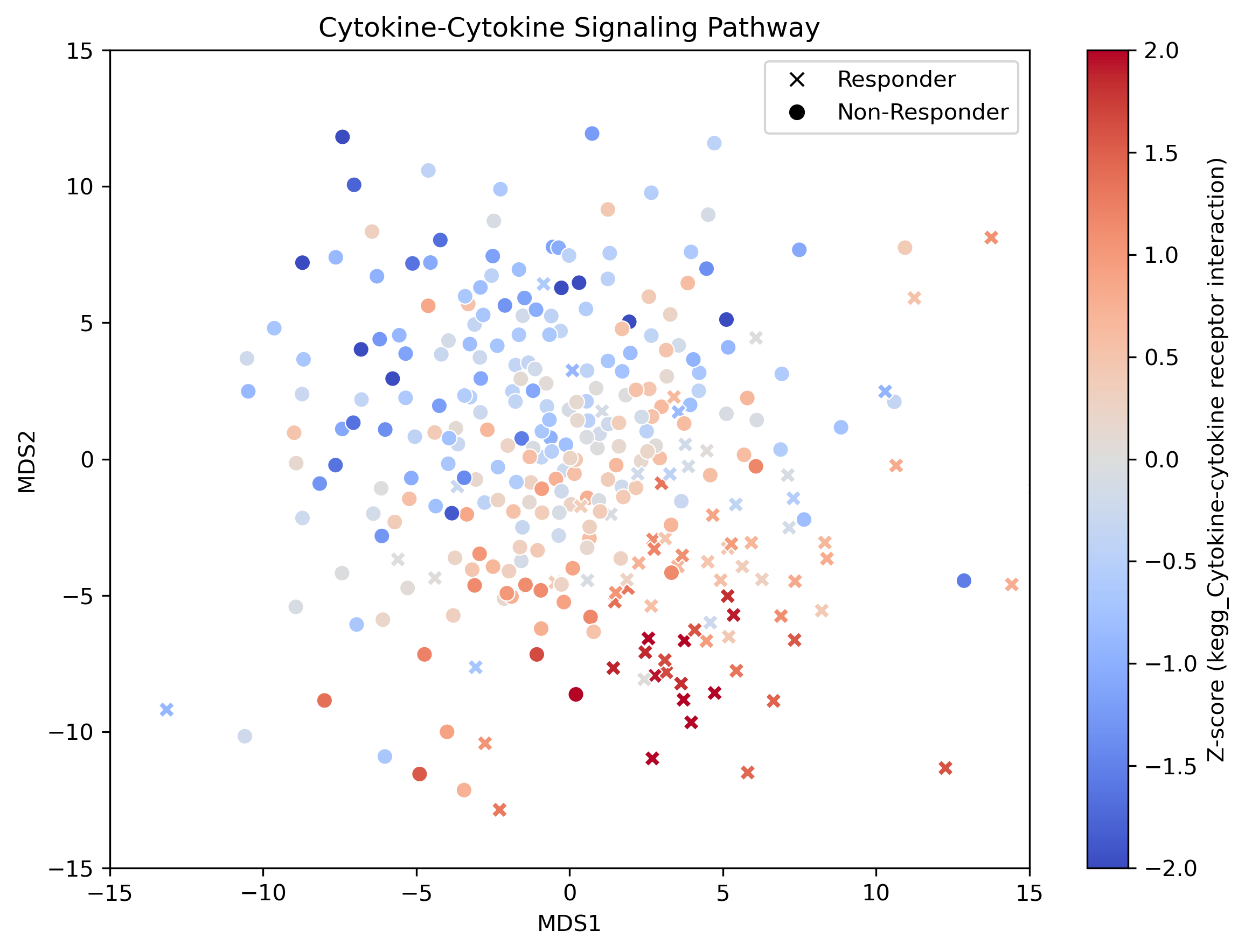}
    \caption{
       MDS projection of BDVAE latent space colored by pathway-level activity for the \textit{Cytokine-Cytokine Receptor Interaction} pathway.
    Samples are colored by standardized z-scores of BDVAE latent activations associated with this pathway.
    Responders (×) and non-responders ($\cdot$) show opposite enrichment patterns, with high activity in non-responders and lower activity in responders.
    C3-Mixed occupies an intermediate region, suggesting a smooth gradient of cytokine-related immune states.
    }
    \label{fig:cytokine_mds_projection}
\end{figure}

At the latent level, MDS projection of BDVAE embeddings colored by pathway-level activity revealed a continuous gradient across the latent space (Figure~\ref{fig:cytokine_mds_projection}). Higher pathway activity was concentrated in the non-responder–enriched region, while responder tumors occupied areas of lower cytokine pathway activation. C3-Mixed was again positioned intermediately, reinforcing its role as a transitional phenotype along the immune-response continuum.

Together, these results identify cytokine dysregulation as a latent axis of resistance to immunotherapy, with a distinct cytokine-attribution gradient that separates responders, non-responders, and molecularly ambiguous tumors.

\subsubsection{Cancer-Associated Signaling Programs Differentiate Clinical Response Groups}

In addition to immune and neuroimmune differences, we examined tumor-intrinsic signaling pathways to evaluate whether oncogenic programs were differentially attributed between responders and non-responders. Supplementary Figure~\ref{fig:cancer_volcano_panels} shows volcano plots of integrated gradients attribution for key cancer-associated pathways, including \textit{Ras Signaling}, \textit{p53 Signaling}, \textit{FoxO Signaling}, \textit{Relaxin Signaling}, \textit{Hippo Signaling}, \textit{Rap1 Signaling}, and \textit{Signaling Pathways Regulating Pluripotency of Stem Cells}.

\begin{figure}[H]
    \centering
    \includegraphics[width=\textwidth]{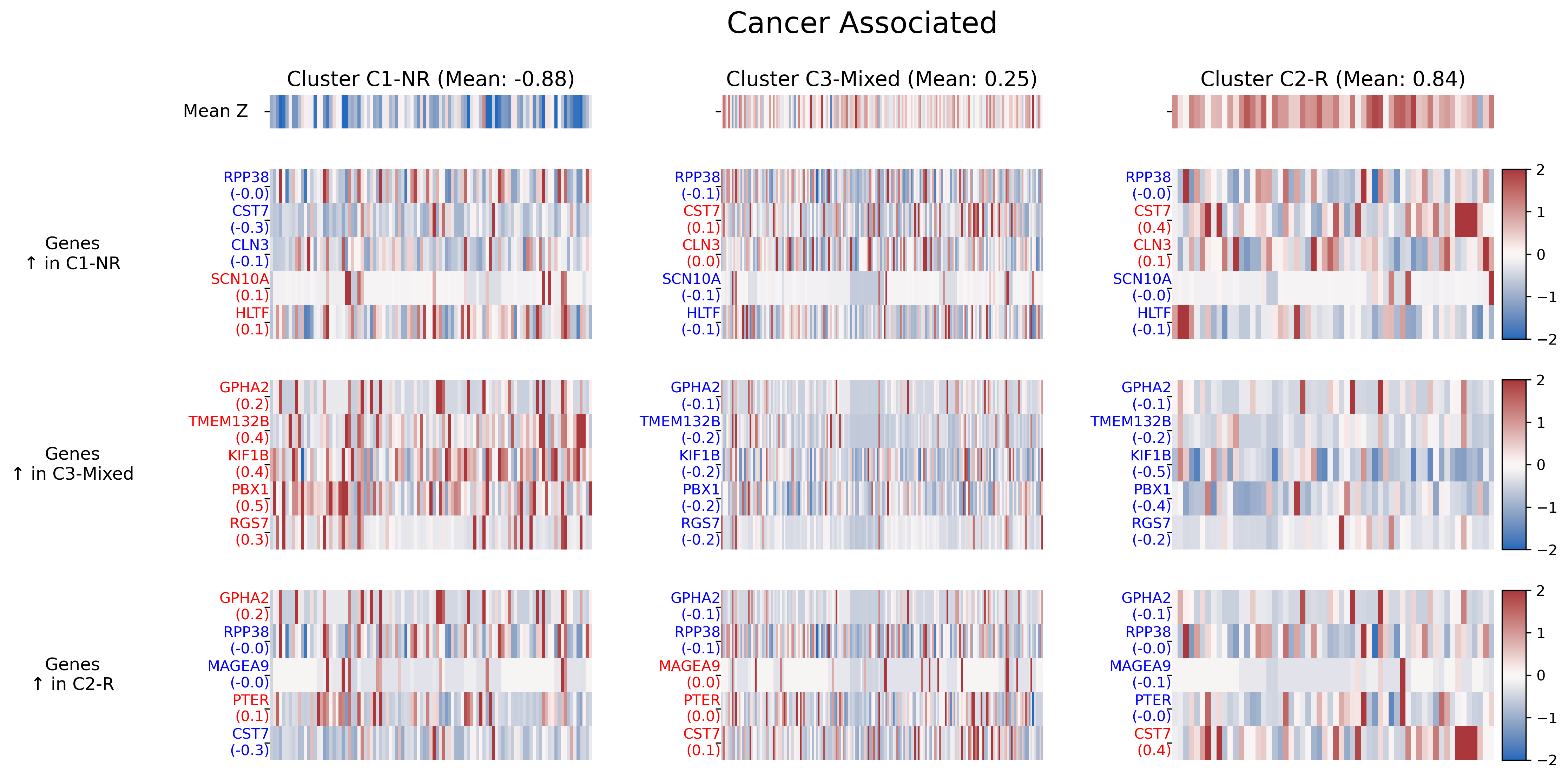}
    \caption{Heatmap of gene expression for top cancer-associated latent features across clusters. Top row shows standardized mean latent activity ($\mathbf{Z}$) for the \textit{Cancer Associated} pathway within each cluster. Rows beneath show expression of genes most strongly attributed to each cluster (C1-NR, C3-Mixed, and C2-R). Red labels indicate positive average expression within the cluster; blue labels indicate negative average expression. Cluster C2-R (responder dominant) shows enrichment for immune-activating and tumor suppressor genes (e.g., \textit{CST7}, \textit{RPP38}, \textit{MAGEA9}), while C1-NR (non-responder dominant) is characterized by genes such as \textit{SCN10A} and \textit{HLTF}, implicated in neuronal and epigenetic regulation. C3-Mixed shows distinct enrichment of developmental and transcriptional regulators (e.g., \textit{PBX1}, \textit{RGS7}, \textit{TMEM132B}), consistent with a transitional state.}
    \label{fig:cancer_associated_heatmap}
\end{figure}

Several distinct patterns emerged, particularly within the “cancer-associated” latent dimensions (Figure~\ref{fig:cancer_associated_heatmap}) as well as the Relaxin and Rap1 signaling pathways.

The Relaxin and Rap1 signaling pathways showed sharp contrasts in latent activity and gene expression across clusters, further highlighting molecular mechanisms underlying clinical response.

\begin{figure}[h]
\centering
\includegraphics[width=\textwidth]{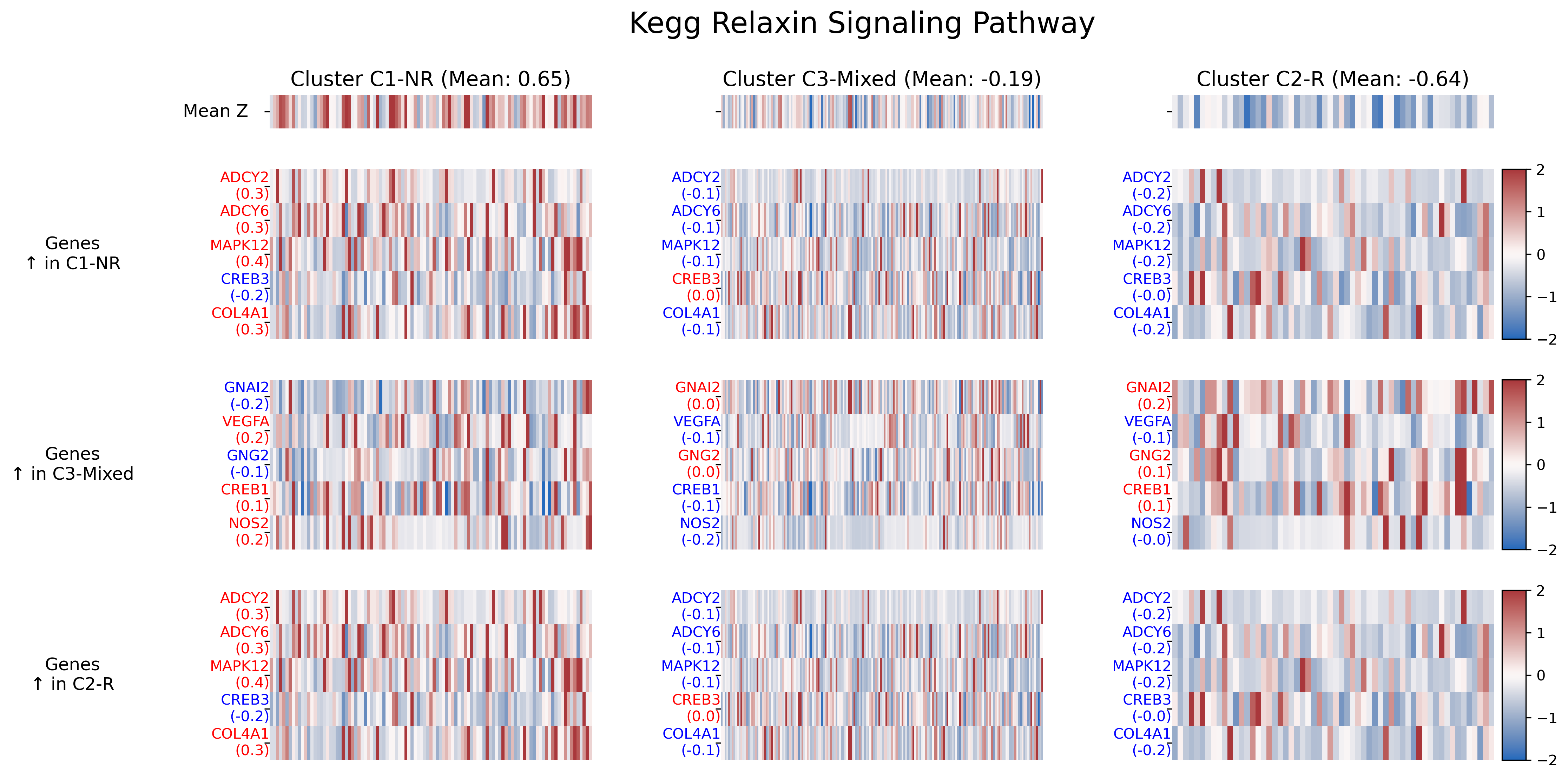}
\caption{
\textbf{Pathway- and gene-level activity in the Relaxin signaling pathway across tumor clusters.}
Top row: mean standardized latent activity (Z-score) derived from BDVAE across clusters. Lower rows show expression of the top genes enriched in each cluster. C1-NR (non-responders) displays elevated latent activity and higher expression of cAMP- and MAPK-associated genes (e.g., \textit{ADCY2}, \textit{ADCY6}, \textit{MAPK12}), suggesting immunosuppressive tumor-intrinsic signaling. C2-R (responders) exhibits suppressed pathway activity. C3-Mixed shows an intermediate profile consistent with a transitional phenotype.
}
\label{fig:relaxin_expression_heatmap}
\end{figure}

Relaxin signaling was most active in the non-responder–dominant cluster (C1-NR), as indicated by high latent scores and upregulation of key components such as \textit{ADCY2}, \textit{ADCY6}, and \textit{MAPK12}. These genes are involved in cAMP production, inflammatory signaling, and extracellular matrix remodeling—features that have been implicated in immune evasion, fibrosis, and altered stromal architecture \cite{Bathgate2013}. In contrast, responders (C2-R) exhibited lower activity and expression in this pathway, suggesting that suppression of relaxin-mediated remodeling may promote immune infiltration or prevent resistance (Figure~\ref{fig:relaxin_expression_heatmap}).

\begin{figure}[h]
\centering
\includegraphics[width=\textwidth]{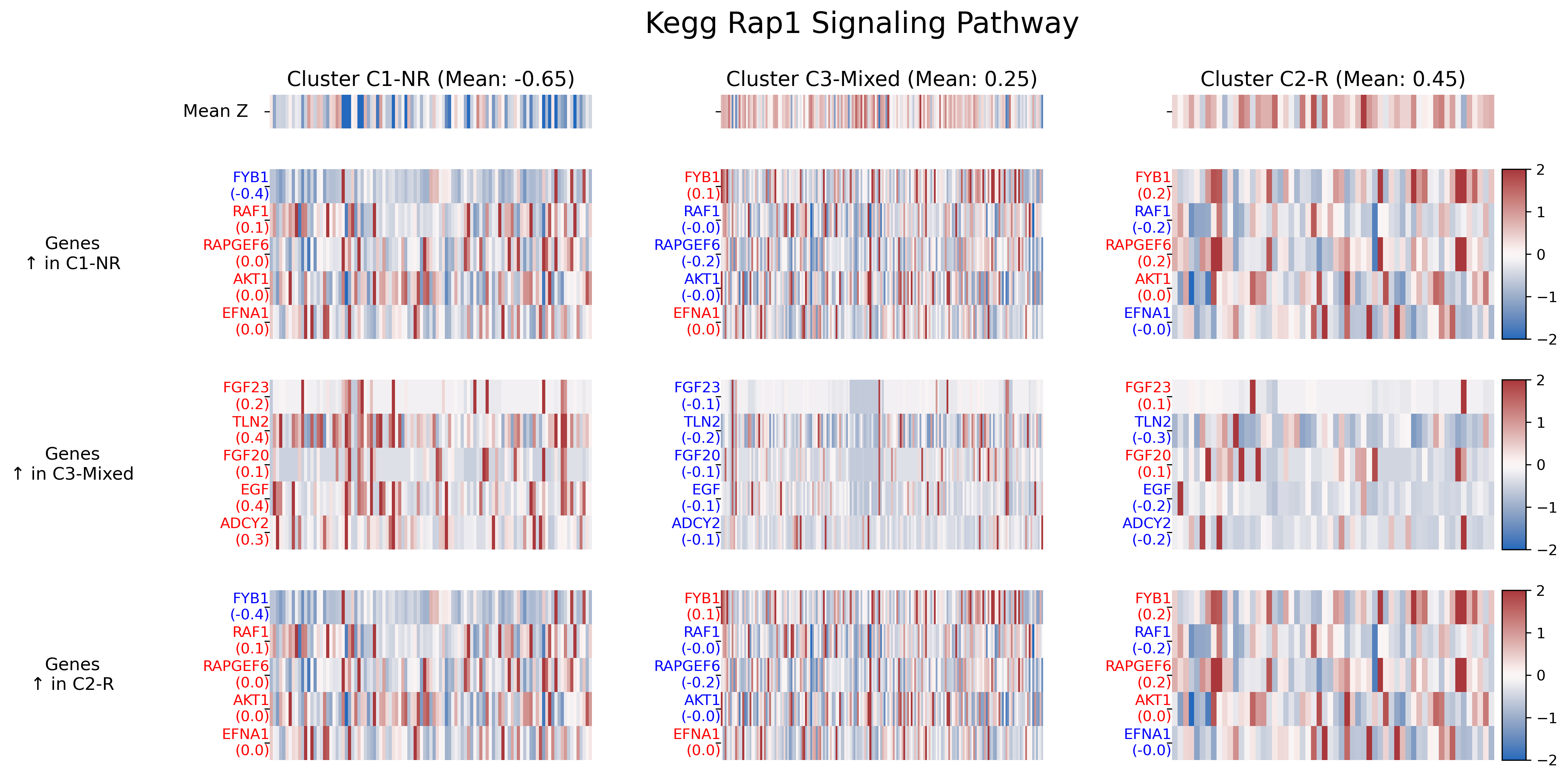}
\caption{
\textbf{Pathway- and gene-level activity in the Rap1 signaling pathway across tumor clusters.}
Responder-enriched cluster C2-R shows high BDVAE-derived latent activity and expression of genes associated with integrin activation and T-cell trafficking (e.g., \textit{RAPGEF6}, \textit{FGF23}, \textit{EGF}). C1-NR (non-responders) exhibits upregulation of tumor-intrinsic survival signals including \textit{RAF1} and \textit{AKT1}. These patterns support a role for Rap1 signaling in effective anti-tumor immunity, and its attenuation in resistance.
}
\label{fig:rap1_heatmap}
\end{figure}

Conversely, the Rap1 signaling pathway showed the opposite trend: responder tumors in C2-R exhibited high latent activation and expression of genes supporting cell adhesion, immune synapse formation, and T-cell trafficking (e.g., \textit{RAPGEF6}, \textit{FGF23}). Rap1 activity has been associated with integrin activation and stable immune synapse formation in cytotoxic T lymphocytes \cite{Katagiri2002}, supporting its potential role in effective anti-tumor immunity. In contrast, non-responders displayed upregulation of \textit{RAF1}, \textit{AKT1}, and other tumor-intrinsic survival signals (Figure~\ref{fig:rap1_heatmap}).

Taken together, these results reveal that tumor-intrinsic signaling programs, in addition to immune-related factors, play a central role in shaping immunotherapy outcomes. Pathways such as Relaxin and Rap1 signaling distinguish responders and non-responders not only through differential latent activation, but also through distinct gene expression profiles aligned with immune-permissive or immune-restrictive microenvironments. The presence of transitional expression patterns in C3-Mixed further suggests that modulation of these oncogenic programs could shift tumors toward a more favorable immune state. Overall, these findings highlight the interpretive power of BDVAE in uncovering clinically meaningful cancer biology, emphasizing the interplay between oncogenic signaling, immune contexture, and therapeutic response.

\subsubsection{Neuroimmune Signaling Differentiates Responders and Non-Responders}

Neuroimmune signaling emerged as a dominant axis of separation between C1-NR and C2-R. Pathways including \textit{Neuroactive Ligand–Receptor Interaction}, \textit{Neurotrophin signaling}, \textit{Retrograde Endocannabinoid signaling}, \textit{GABAergic cAMP signaling}, and \textit{Calcium signaling} showed strong enrichment patterns across clusters. 

To pinpoint pathway components driving these distinctions, we examined gene-level attributions from integrated gradients (IG). For each gene, Kruskal–Wallis tests were used to assess differences in IG attribution across clusters (C1-NR, C2-R, C3-Mixed), followed by FDR correction. Effect sizes were further quantified using Cliff’s delta. Genes with significant attribution differences (FDR $< 0.05$) and large effects ($|\delta| > 0.33$) were prioritized. This analysis highlighted discrete sets of neuroimmune genes that stratify responders and non-responders, suggesting that dysregulated neuroimmune cross-talk underpins differential therapeutic outcomes.

A focused view of neuronal and neuroimmune signaling is provided in Supplementary Figure~\ref{fig:neuro_pathway_volcano_grid}. 

Notably, three of the most differentially attributed pathways—\textit{Calcium signaling}, \textit{cAMP signaling}, and \textit{Neuroactive Ligand–Receptor Interaction}—are integral to neuroimmune regulation and intersect with T-cell activation and tumor–immune interface signaling. The elevated model attribution of features from these pathways in the non-responder–dominant cluster suggests that resistance to immunotherapy may involve neuroimmune mechanisms such as disrupted calcium flux, altered cAMP signaling tone, or the presence of immunoinhibitory neurotransmitter receptors. While these patterns reflect model-derived feature importance rather than differential gene expression, they are consistent with prior reports implicating neuroimmune suppression in tumor immune evasion \cite{Jin2013, Dionisio2011, Bhat2009}.

Within the \textit{Neuroactive Ligand–Receptor Interaction} pathway, features receiving higher IG attribution in the C2-R cluster (responder dominant) included genes associated with cytotoxicity, neuroinflammatory signaling, and immune co-stimulation, such as \textit{GZMA}, \textit{P2RY6}, \textit{GRIN2A}, \textit{GLP1R}, and \textit{P2RX7}. These genes have established roles in T-cell effector function, pro-inflammatory signaling, and neuroimmune communication \cite{Lieberman2003, Moore2003, Adinolfi2015, Takata2021, Li2022}.

Conversely, in the C1-NR cluster (non-responder dominant), genes contributing more strongly to latent feature attributions included GABAergic inhibitory receptor subunits such as \textit{GABBR1}, \textit{GABRP}, \textit{GABRB2}, and \textit{GABRA4}. These genes are known to regulate neuronal excitability and have been implicated in immunosuppressive signaling within the tumor microenvironment \cite{Jin2013, Wheeler2011, Dionisio2011}.

These attribution patterns suggest that the BDVAE model distinguishes clinical response groups, in part, based on divergent neuroimmune signaling logic—highlighting pro-inflammatory and immune-stimulatory components in responders and immunoinhibitory features in non-responders. This finding underscores neuroimmune signaling as a potentially targetable mechanism in resistance to immune checkpoint therapy.

\begin{figure}[H]
    \centering
    \includegraphics[width=0.95\textwidth]{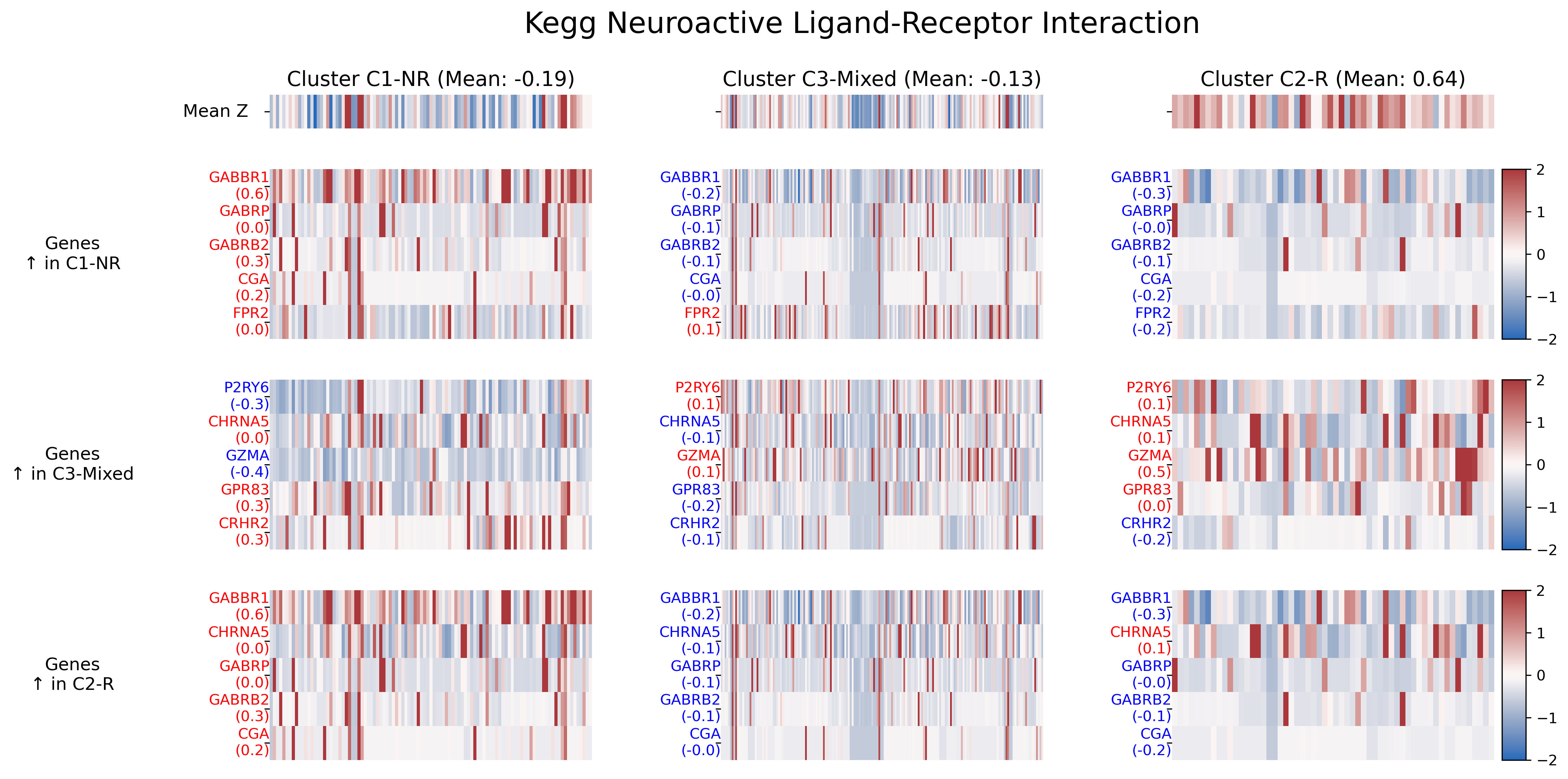}
    \caption{
    Heatmap summarizing neuroimmune variation in the \textit{Neuroactive Ligand-Receptor Interaction} pathway across tumor clusters. 
    The top row shows the mean pathway-level latent activity (Z score) derived from BDVAE, averaged across all significant latent variables associated with this pathway. 
    Lower panels show standardized gene expression values for the top 5 genes that contribute most strongly to pathway-level attribution in each cluster. 
    Columns correspond to individual tumor samples, grouped by cluster (C1-NR, C3-Mixed, C2-R). }
    \label{fig:neuroactive_expression_heatmap}
\end{figure}

To further interpret this axis, we examined gene expression patterns for selected components of the \textit{Neuroactive Ligand–Receptor Interaction} pathway across clusters. As shown in Figure~\ref{fig:neuroactive_expression_heatmap}, responder-enriched tumors in C2-R displayed higher expression of immune-activating and cytotoxic genes, including \textit{GZMA}, \textit{P2RY6}, and \textit{CHRNA5}, which are associated with granzyme-mediated cytolysis, ATP-sensing pro-inflammatory signaling, and neuronal co-stimulation of immune responses.

Interestingly, several genes that contributed most strongly to the latent structure of C2-R—such as \textit{GABBR1}, \textit{GABRP}, \textit{GABRB2}, and \textit{CGA}—and C3-Mixed—such as \textit{TAC1}, \textit{GAL}, and \textit{C5AR1}—exhibited a consistent expression gradient across clusters. For these genes, expression was highest in non-responders (C1-NR), lowest in responders (C2-R), and intermediate in C3-Mixed. This pattern reinforces the interpretation of C3-Mixed as a biologically heterogeneous or transitional immune state, potentially reflecting tumors in flux between immunosuppressive and immunoreactive phenotypes.

\begin{figure}[H]
    \centering
    \includegraphics[width=0.6\textwidth]{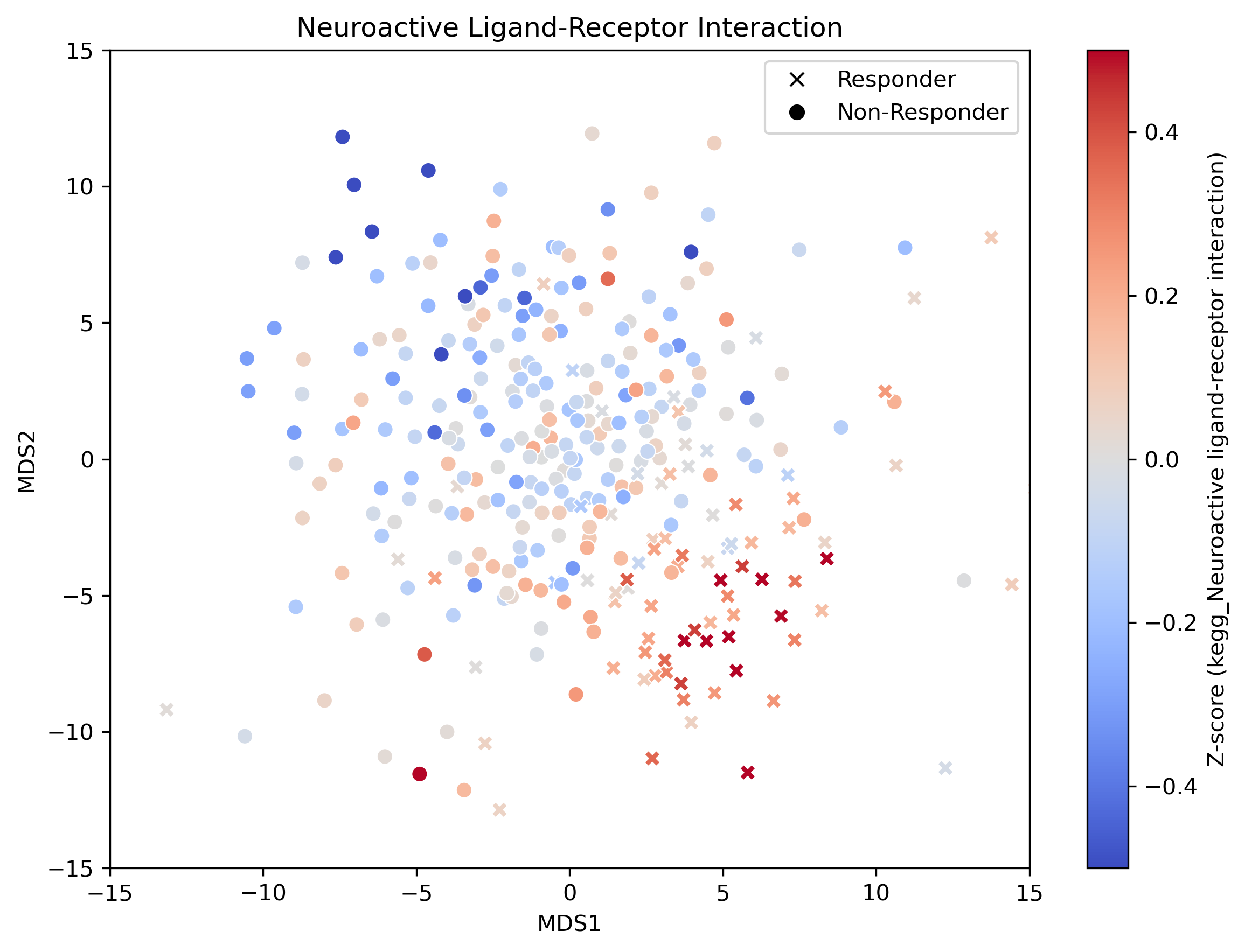}
    \caption{
        Multidimensional scaling (MDS) projection of tumor samples in BDVAE latent space, colored by standardized pathway-level activity for the \textit{Neuroactive Ligand-Receptor Interaction} pathway. 
        Pathway activity was computed as the mean z-scored latent activation for all BDVAE features annotated to this pathway. 
        Responders (×) and non-responders ($\cdot$) are overlaid to show alignment with clinical outcome. 
        A gradient is evident, with higher pathway activity (red) concentrated in responder-enriched regions of latent space, and lower activity (blue) more common among non-responders. 
        This spatial organization suggests that BDVAE captures a latent neuroimmune axis associated with therapeutic response.
    }
    \label{fig:neuroactive_mds_projection}
\end{figure}

To visualize how BDVAE captures this neuroimmune axis at the latent level, we projected tumor samples into a two-dimensional MDS embedding of the latent space (Figure~\ref{fig:neuroactive_mds_projection}). Samples were colored by standardized pathway-level activity for the \textit{Neuroactive Ligand–Receptor Interaction} pathway. A clear gradient emerged: tumors in the responder-enriched region of latent space displayed elevated pathway activity, while those in non-responder–dominated regions exhibited lower values. This pattern reinforces the finding that BDVAE captures neuroimmune signaling as a meaningful latent dimension underlying immunotherapy response.

\subsubsection{Metabolic and Miscellaneous Signaling Features Underpin Clinical Variation}

Beyond immune, neuroimmune, and oncogenic axes, we identified differential attribution across several metabolic and context-specific signaling pathways, revealing additional biological mechanisms that may underlie tumor cell state, immune modulation, and therapeutic resistance.

Latent dimensions associated with hormonal regulation (e.g., insulin and glucagon signaling), oxidative stress, and tissue remodeling emerged as significantly differentiated between clinical clusters. Notably:
\begin{itemize}
    \item \textbf{C1-NR (non-responders)} showed elevated attribution for anabolic and pro-survival signaling programs, including genes such as \textit{IRS1}, \textit{AKT1}, \textit{PRKAA2}, and \textit{PLCB2}, suggesting an immune-exclusion phenotype maintained through metabolic dominance.
    \item \textbf{C2-R (responders)} exhibited increased latent activity and expression of catabolic and oxidative stress–related genes such as \textit{FOXO1}, \textit{PGAM2}, and \textit{ADCY2}, suggesting a more metabolically vulnerable or immune-permissive state.
    \item \textbf{C3-Mixed} demonstrated intermediate patterns, potentially reflecting transitional metabolic plasticity.
\end{itemize}

\begin{figure}[H]
\centering
\includegraphics[width=\textwidth]{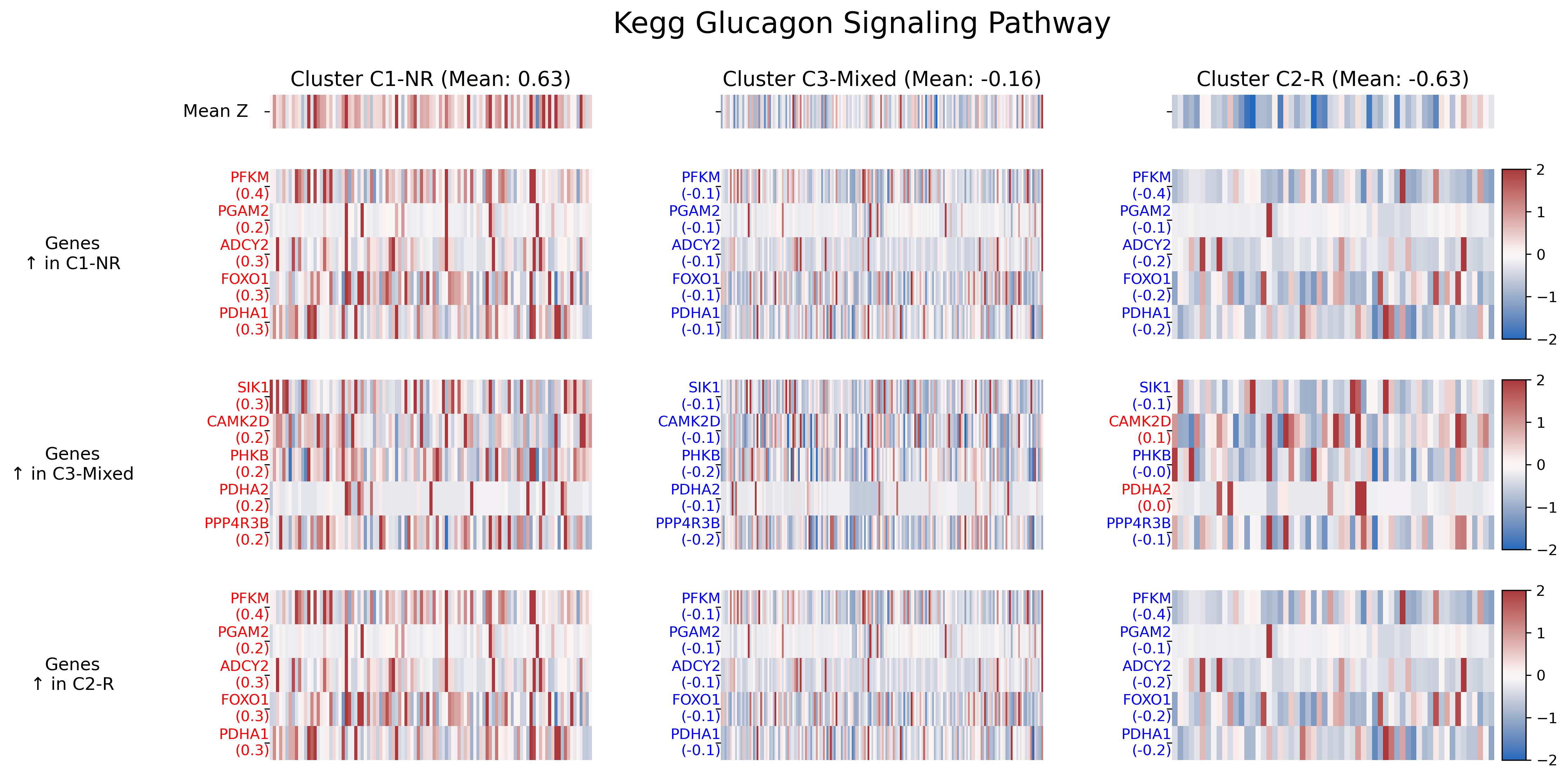}
\caption{
\textbf{Latent activity and gene expression in the Glucagon signaling pathway across clusters.}
The top row shows standardized latent activity ($\mathbf{Z}$-scores) for the Glucagon signaling pathway in each cluster. C2-R (responders) displays elevated expression of genes such as \textit{FOXO1}, \textit{PGAM2}, and \textit{ADCY2}, which are involved in glucose metabolism, oxidative stress resistance, and cAMP signaling. In contrast, non-responders (C1-NR) show upregulation of \textit{PRKAA2} and \textit{PLCB2}, consistent with altered energy sensing and phospholipid signaling. These patterns suggest metabolic divergence across response groups, potentially reflecting differences in tumor adaptability or immune–metabolic crosstalk.
}
\label{fig:glucagon_signaling_heatmap}
\end{figure}

\begin{figure}[H]
\centering
\includegraphics[width=\textwidth]{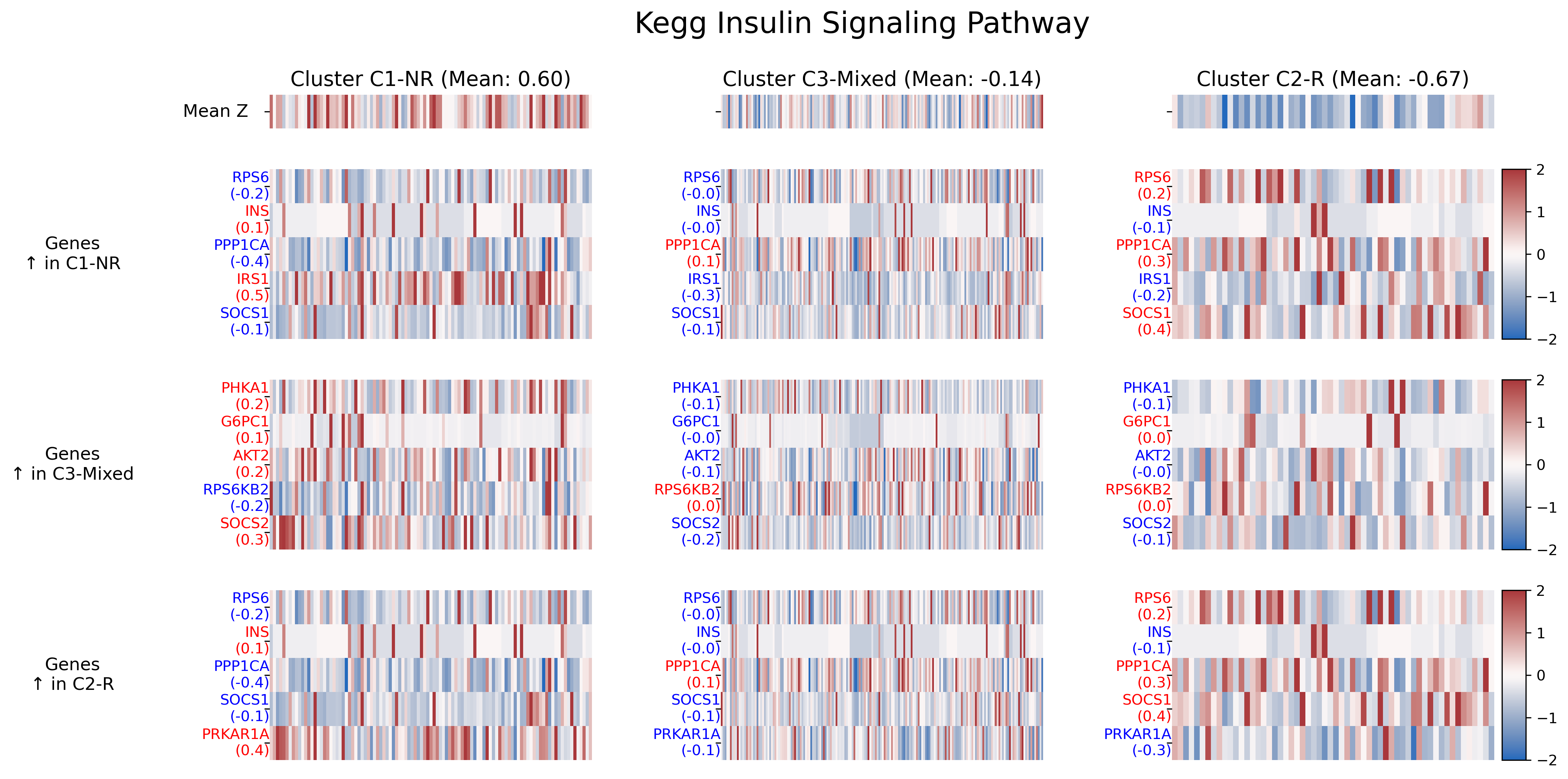}
\caption{
\textbf{Latent activity and gene expression of the Insulin signaling pathway across clusters.}
C1-NR tumors exhibit higher latent activity and expression of metabolic and survival-associated genes including \textit{IRS1}, \textit{AKT1}, and \textit{PPP1CA}. These features suggest an anabolic, insulin-responsive phenotype in non-responders. Conversely, responder tumors (C2-R) show suppression of these signatures, potentially reflecting metabolic reprogramming or immune-driven catabolism. C3-Mixed exhibits intermediate expression patterns.
}
\label{fig:insulin_signaling_heatmap}
\end{figure}

\begin{figure}[H]
\centering
\includegraphics[width=\textwidth]{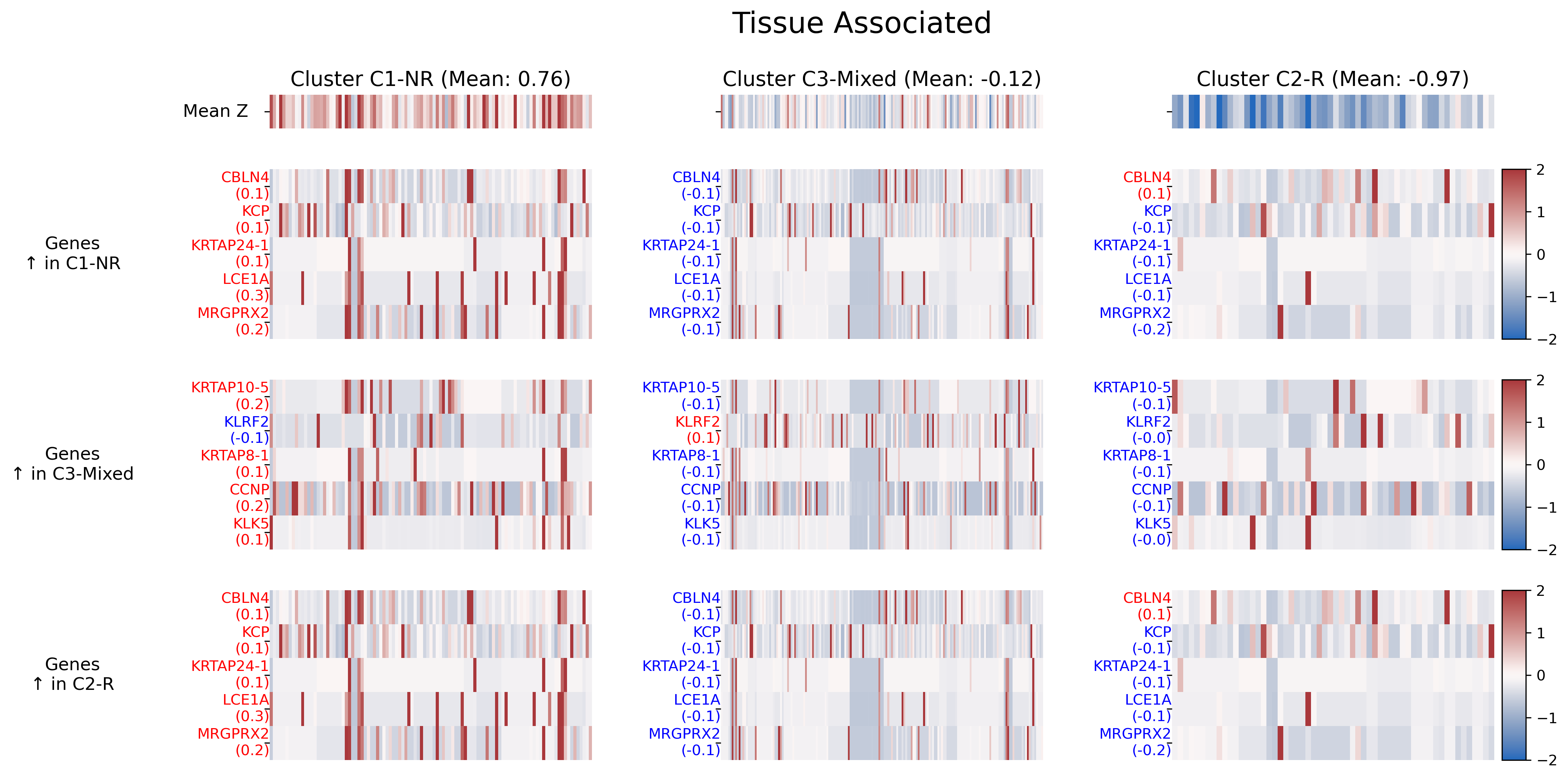}
\caption{
\textbf{Latent and gene-level profiles for tissue-associated features across tumor clusters.}
The top row shows standardized BDVAE-derived latent activity for the "Tissue Associated" gene set. Lower panels display expression of genes most strongly attributed to each cluster. Non-responders (C1-NR) exhibit elevated expression of tissue-remodeling and keratinocyte-associated genes (e.g., \textit{KRTAP24-1}, \textit{LCE1A}, \textit{MRGPRX2}), consistent with stromal remodeling and barrier-tissue phenotypes. In contrast, responder-enriched tumors (C2-R) show marked suppression of these programs, suggesting a more immune-permissive microenvironment.
}
\label{fig:tissue_associated_heatmap}
\end{figure}

These findings suggest that non-responder tumors exhibit features of metabolic rigidity—characterized by sustained expression of anabolic and pro-survival genes such as \textit{IRS1}, \textit{AKT1}, \textit{PRKAA2}, and \textit{PLCB2}. This transcriptional profile is consistent with a stable, insulin-responsive metabolic state that may resist adaptive reprogramming under immunologic stress. In contrast, responder-enriched tumors demonstrate greater engagement of stress-responsive and catabolic pathways (e.g., \textit{FOXO1}, \textit{PGAM2}), suggesting a capacity for metabolic flexibility. The absence of such adaptive signatures in non-responders, along with their lower transcriptional heterogeneity, supports the hypothesis that metabolic rigidity may limit immune infiltration or impair effective anti-tumor immunity. These interpretations are supported by prior evidence linking metabolic competition and inflexible energy programs to immune exclusion and checkpoint inhibitor resistance \cite{Chang2015, Ho2015, Leone2019, Scharping2021}.

\subsubsection{Quantifying Label-Outcome Discordance via Progression-Free Survival}

While RECIST-defined response labels provide a standard framework for evaluating treatment benefit, we observed that some patients exhibited PFS patterns inconsistent with their response classification. Specifically, a subset of responders had relatively short PFS, while some non-responders experienced unusually prolonged PFS. These “misaligned” cases—patients whose survival outcomes diverge from typical expectations based on RECIST—raise the possibility of underlying biological heterogeneity not captured by conventional RECIST criteria. To explore this further, we systematically identified and characterized misaligned samples using quantile-based thresholds, aiming to uncover molecular features associated with discordant clinical trajectories. 

To identify discordant cases between RECIST-defined response and actual progression outcomes, we defined \textit{misaligned samples} based on extreme progression-free survival (PFS) within each response category. Specifically, \textit{misaligned responders} were defined as patients labeled as responders by RECIST criteria whose PFS fell below the 5th percentile of the responder distribution, and \textit{misaligned non-responders} were defined as RECIST non-responders whose PFS exceeded the 95th percentile of the non-responder distribution. This quantile-based approach isolates samples whose survival trajectories deviate markedly from their assigned RECIST class, enabling downstream analyses of biological or model-inferred features that may underlie this discordance. 

To statistically validate the distinction between \textit{misaligned} and \textit{aligned} samples within each response group, we compared their PFS distributions using two-sided Mann–Whitney U tests. Misaligned non-responders had significantly longer PFS than the rest of the non-responder group (U = 684.00, p = 9.44 × 10\textsuperscript{-5}), while misaligned responders had significantly shorter PFS than other responders (U = 2.00, p = 0.0272). These findings confirm that the identified misaligned samples are indeed outliers within their respective groups in terms of time to progression, further justifying their analysis as biologically distinct or misclassified cases.  

\section{Discussion and Conclusion}

We present a Biologically Disentangled Variational Autoencoder (BDVAE) for uncovering mechanisms of differential response to immune checkpoint blockade (ICB). By jointly modeling RNA-seq and whole-exome sequencing data across four cancer types, BDVAE learns low-dimensional, interpretable representations that capture clinically relevant tumor states. The model distinguishes responders from non-responders with high predictive accuracy, while also revealing intermediate phenotypes that conventional approaches overlook.

Interpretation of latent dimensions highlighted immune-regulatory, oncogenic, metabolic, and neuroimmune programs as key axes of variation. Clustering of latent features identified three biologically coherent groups: a responder-enriched cluster, a non-responder–enriched cluster, and a heterogeneous intermediate cluster. The intermediate group exhibited mixed immune and tumor-intrinsic signatures, suggesting a transitional phenotype that may be particularly responsive to combination strategies. Validation against progression-free survival confirmed that BDVAE-derived clusters stratify patients along clinically meaningful trajectories, even in cases misaligned with RECIST labels. These findings indicate that BDVAE captures heterogeneity beyond standard response definitions, identifying subgroups with distinct biology and therapeutic potential.

The framework provides a biologically structured lens for patient stratification and biomarker discovery. Pathway-level signals—including Rap1, Relaxin, and metabolic programs linked to immune exclusion—nominate potential therapeutic targets, while neuroimmune signaling pathways highlight underappreciated intersections between nervous and immune regulation in cancer. Gene-level analyses further suggest candidate drivers of resistance or sensitivity, including \textit{JAK1}, \textit{STAT3}, \textit{MAP2K1}, and \textit{NRAS}. Together, these insights illustrate how disentangled representation learning can bridge predictive modeling and mechanistic interpretation in precision oncology.

Several limitations warrant mention. The dataset, while spanning multiple cancer types, remains modest in size and pathway annotations do not include all possible pathways which may lead to an incomplete picture. Moreover, spatial and longitudinal dimensions of tumor-immune interactions are not captured. Future work should expand BDVAE to incorporating single-cell, spatial, and imaging data, as well as longitudinal trajectories of response and resistance. Prospective validation of BDVAE-derived biomarkers in clinical trials, alongside experimental testing of implicated pathways, will be critical to translating these findings into therapeutic strategies.

In summary, BDVAE demonstrates how interpretable generative models can recover established and novel mechanisms of ICB response, revealing discrete phenotypes as well as potential transition states. By integrating predictive performance with biological insight, this approach provides a foundation for precision stratification and therapeutic hypothesis generation in cancer immunotherapy.

\section{Methods}
\subsection{Data Pre-Processing}
\subsection*{Public and Controlled Data Acquisition}

Publicly available raw RNA-seq and whole-exome sequencing (WES) datasets were obtained directly from the Sequence Read Archive (SRA). Access to controlled clinical datasets required approval through the Database of Genotypes and Phenotypes (dbGaP) or the European Genome-phenome Archive (EGA), following the respective data use policies. Datasets summarized in Table~\ref{tab:minimal_study_summary}.

\begin{table}[H]
\centering
\caption{Summary of included studies and datasets used in this work.}
\renewcommand{\arraystretch}{1.2}
\begin{tabular}{|l|l|p{6cm}|}
\hline
\textbf{Study} & \textbf{Cancer Type} & \textbf{Accession ID(s)} \\
\hline
McDermott et al., 2019 \cite{mcdermott2019atezolizumab} & Renal Cell Carcinoma & EGAD00001004183 \\
\hline
Mariathasan et al., 2018 \cite{mariathasan2018tgf} & Urothelial Cancer & EGAD00001003977; EGAD00001004218 \\
\hline
Riaz et al., 2017 \cite{riaz2017melanoma} & Melanoma & SRP095809; SRP094781 \\
\hline
Hugo et al., 2016 \cite{hugo2016genomic} & Melanoma & SRP067938; SRP090294 \\
\hline
Kim et al., 2018 \cite{kim2018pembrolizumab} & Gastric Cancer & PRJEB25780 \\
\hline
Van Allen et al., 2015 \cite{vanallen2015genomic} & Melanoma & phs000452 \\
\hline
\end{tabular}
\label{tab:minimal_study_summary}
\end{table}

\subsection*{RNA-Sequencing Processing}
Raw FASTQ files were processed as follows:
\begin{enumerate}
    \item Adapter trimming with \texttt{fastp} (v0.23.2).
    \item Alignment to GRCh38 with \texttt{STAR} (v2.7.10a).
    \item Gene- and transcript-level quantification with \texttt{RSEM} (v1.3.3).
    \item Quality assessment with \texttt{RNA-SeQC2} (v2.4.2).
    \item Exclusion of libraries with $<$10M mapped reads, duplication rates $>$60\%, or median 3’ bias $>$20\%.
    \item Gene-level expression matrices were TPM-normalized, log$_2$(TPM+1) transformed, and z-score standardized.
\end{enumerate}
The resulting expression matrices were used as RNA-seq features. 

\subsection*{Whole Exome Sequencing (WES) Processing}
Whole-exome BAMs were processed according to GATK Best Practices:
\begin{enumerate}
    \item Alignment with \texttt{BWA-MEM} (v0.7.17).
    \item Duplicate marking with \texttt{Picard} (v2.25.0).
    \item Base quality score recalibration with \texttt{GATK} (v4.2.6).
    \item Somatic variant calling with \texttt{Mutect2}.
    \item Filtering with \texttt{FilterMutectCalls}, \texttt{GetPileupSummaries}, and \texttt{CalculateContamination}.
    \item Annotation with \texttt{Funcotator} and \texttt{OpenCravat}.
\end{enumerate}
Annotated variant call files were converted to MAF format for downstream analyses. 

\subsection*{Somatic and Derived Features}
From annotated MAFs, we extracted:
\begin{itemize}
    \item \textbf{Tumor Mutational Burden (TMB):} Non-synonymous SNVs per Mb.
    \item \textbf{Mutational Signatures:} SBS features using COSMIC v3 reference profiles.
    \item \textbf{CADD Features:} Raw and PHRED-scaled CADD annotations aggregated at the sample level.
    \item \textbf{Oncogenic Summary Metrics:} Derived via \texttt{maftools}, including counts of oncogenic drivers and druggable variants.
\end{itemize}

\subsection*{Clinical Harmonization}
RECIST categories, censoring rules, and PFS definitions were standardized across cohorts. Table~S2 reports harmonized clinical covariates.

\paragraph{Batch correction.} 
When metadata contained cohort or source labels, we applied empirical Bayes correction with \texttt{ComBat} (\texttt{scanpy.pp.combat}) to adjust for study-specific effects. This procedure was performed using cohort labels as the batch covariate, and ensured that downstream models learned signal attributable to biological and clinical variation rather than technical batch effects.

\subsection*{Integration and Feature Standardization}
RNA-seq features, WES-derived features, and clinical labels were merged at the sample level. Continuous features were z-score standardized, while sparse binary features were retained in raw form. Samples missing any modality were excluded, yielding a final cohort of 366 patients across four tumor types.

\subsection*{Encoder Definitions}
The Biologically Disentangled Variational Autoencoder (BDVAE) incorporated structured encoders derived from curated biological feature sets, constraining latent variables to align with interpretable processes.

\subsubsection*{Immune Signatures}
Immune-related gene sets were curated from the Tumor Immune Dysfunction and Exclusion (TIDE) framework and related literature \cite{Fu2020}. Each encoder corresponded to a distinct immune cell state or functional program, with positive and negative variants capturing directionality (e.g., T cell exhaustion, T cell accumulation). A full list is provided in Supplementary Table~\ref{tab:supp_tide_encoders}.

\subsubsection*{Tissue-Specific Signatures}
Tissue-enriched gene sets were obtained from the Human Protein Atlas (HPA) and GTEx (v8, 2023). Genes annotated as ``tissue enriched’’ or ``tissue enhanced’’ were grouped into encoders (e.g., adrenal, kidney, lung, skin). These were merged into a composite encoder labeled \texttt{tissue-associated}.

\subsubsection*{Cancer-Associated Signatures}
Tumor-type-specific gene sets were compiled from COSMIC, DisGeNET, IntOGen, and MalaCards. Genes were filtered using organ- or tumor-specific keywords (e.g., ``melanoma,’’ ``renal cell carcinoma’’). The merged encoders were labeled \texttt{cancer-associated} (Supplementary Table~\ref{tab:supp_cancer_encoders}).

\subsubsection*{WES-Derived Features}
Feature groups were derived using \texttt{maftools}, including drug-response indicators, SBS signatures, and oncogenic burden metrics. Variants were annotated with CADD scores, and the maximum PHRED-scaled score per sample was retained (Supplementary Table~\ref{tab:supp_wes_features}).

\subsubsection*{CADD-Based Annotations}

To quantify the functional impact of somatic variants at base-level resolution, we incorporated annotations from the Combined Annotation Dependent Depletion (CADD) framework. CADD integrates multiple layers of annotation—ranging from evolutionary conservation to regulatory elements and coding potential—into a single model that scores the deleteriousness of variants genome-wide.

Somatic variants identified from WES were annotated with CADD using the following steps:

\begin{enumerate}
    \item \textbf{Variant Calling:} Somatic single-nucleotide variants (SNVs) were obtained using the preprocessing workflow described previously, and exported in VCF format for downstream annotation.
    
    \item \textbf{Annotation with CADD:} We annotated each variant with the complete set of raw CADD features. Rather than using only the final PHRED-scaled score, we extracted all available annotations from the \texttt{cadd.tsv.gz} output.
       
    \item \textbf{Sample-Level Aggregation:} Since CADD annotations are computed at the variant level, we aggregated them by selecting the maximum PHRED-scaled CADD score across all variants detected in each sample.
\end{enumerate}

These features enable the model to capture fine-grained mutational impact signatures and provide orthogonal information beyond canonical mutation burden metrics.

\subsubsection*{Pathway-Level Features}
We retrieved curated pathway gene sets from the \texttt{KEGG\_2021\_Human} library using the GSEApy interface. Each KEGG pathway was mapped to its associated gene list, and we computed the degree of overlap with both RNA-seq and WES-derived feature sets. Only pathways with at least 10 genes were considered.

Latent features prefixed with \texttt{wes\_} contained the PHRED-scaled CADD pathogenicity scores for each gene, while latent features associated with RNA-Seq data contained normalized gene expression data. 

\begin{enumerate}
    \item \textbf{Loading and Preprocessing:} We converted the KEGG gene sets to a pandas DataFrame containing the pathway name, gene list, and total gene count.
    
    \item \textbf{RNA-seq Feature Overlap:} For each pathway, we calculated the fraction of genes that overlap with the RNA-seq gene expression matrix:
    \[
    \texttt{total\_overlap\_rna} = \frac{|\texttt{genes} \cap \texttt{rna\_features}|}{|\texttt{genes}|}
    \]
    
    \item \textbf{WES Feature Overlap via CADD:} To compute overlap with WES-derived features, we converted each gene name into a prefixed CADD feature (e.g., \texttt{cadd\_TP53}) and computed:
    \[
    \texttt{total\_overlap\_wes} = \frac{|\texttt{cadd\_genes} \cap \texttt{wes\_features}|}{|\texttt{genes}|}
    \]
    
    \item \textbf{Pathway Filtering:} We retained KEGG pathways whose names contained one of the following terms: \texttt{``pathway''}, \texttt{``interaction''}, or \texttt{``transporter''}. These were selected as they represent biologically coordinated systems relevant to tumor-immune interactions.
    
    \item \textbf{Dictionary Construction:} The filtered pathways were prefixed with \texttt{kegg\_} and compiled into a dictionary:
    \[
    \texttt{kegg\_main\_dict[pathway\_name]} \rightarrow \texttt{gene\_list}
    \]
\end{enumerate}

This step enabled integration of curated pathway-level features into the BDVAE model, allowing for structured biological interpretation of latent factors.

\subsection{Model Architecture}

We implemented a \textbf{Biologically Disentangled Variational Autoencoder (BDVAE)} in PyTorch Lightning (v2.6.0+cu124) to learn modular latent representations from high-dimensional omics profiles ($10{,}659$ features per sample). The model consists of three major components: The encoders, decoder, and the supervised classifier head. In total, the architecture contains $\sim 84.6$M trainable parameters.

\subsection*{Encoder} 
The encoder (\texttt{BDEncoder}, $\sim 2.5$M parameters) is factorized into a bank of 321 EncoderFactors, each of which receives as input a masked subset of the original features. Within each factor, masked inputs are passed through a Linear transformation followed by a LeakyReLU activation, and then mapped to the parameters of a Gaussian distribution $(\mu, \sigma)$. Latent variables are obtained using the reparameterization trick. 

Each EncoderFactor is assigned a fixed number of latent dimensions. Two strategies are supported. In the first, the user specifies the overall latent dimensionality, and each EncoderFactor is allocated a number of latent features proportional to the size of its masked input subset. In the second, the dimensionality of each EncoderFactor is automatically determined using a data-driven heuristic: principal component analysis (PCA) is applied to the masked feature subset, and the elbow point of the explained variance curve is identified with the KneeLocator algorithm. The resulting elbow point defines the number of latent dimensions assigned to that EncoderFactor.  

Concatenation of all EncoderFactor outputs yields a latent representation of approximately $2000$ dimensions in the best performing model following hyperparameter optimization, which is then used by both the decoder and classifier.

\subsection*{Decoder} 
The decoder ($\sim 80.1$M parameters) reconstructs the original feature space from the latent representation. It is implemented as a two-layer fully connected network: 
\[
z \;\rightarrow\; \texttt{Linear} \;\rightarrow\; \texttt{LeakyReLU} \;\rightarrow\; \texttt{Linear} \;\rightarrow\; \texttt{LeakyReLU} \;\rightarrow\; \texttt{Softplus} \;\rightarrow\; \texttt{Sigmoid}.
\]
This architecture ensures smooth outputs constrained to the range $[0,1]$. The decoder accounts for the majority of the model’s parameters.

\subsection*{Classifier} 
A supervised classification head ($\sim 2.0$M parameters) is attached to the latent space to predict binary treatment response. The classifier consists of:
\[
\texttt{Linear}(2000 \rightarrow 1000) \;\rightarrow\; \texttt{LeakyReLU} \;\rightarrow\; \texttt{Linear}(1000 \rightarrow 1),
\]
producing a single logit output.

\subsection{Training Objective}

The model was trained to optimize reconstruction, regularization, and supervised prediction objectives jointly:
\begin{itemize}
    \item \textbf{Reconstruction loss:} mean squared error between inputs $x$ and reconstructions $\hat{x}$.
    \item \textbf{Regularization loss:} maximum mean discrepancy (MMD) between the posterior $q(z|x)$ and an isotropic Gaussian prior $p(z)$.
    \item \textbf{Supervised loss:} binary cross-entropy (BCE) applied to classifier logits.
\end{itemize}

The total loss is defined as:
\[
\mathcal{L} = \underbrace{\|x - \hat{x}\|^2}_{\text{Reconstruction}} \;+\; \underbrace{\text{MMD}(z, p(z))}_{\text{Regularization}} \;+\; \beta \cdot \underbrace{\text{BCE}(y, \hat{y})}_{\text{Supervised}} ,
\]
where $\beta$ is a weighting hyperparameter.

\subsection{Training Procedure}

All models were implemented in \texttt{PyTorch Lightning}, with reproducibility enforced by seeding Python, NumPy, and PyTorch random number generators.

\subsection*{Optimization} 
Training was performed with manual optimization using two AdamW optimizers: one for the encoder and decoder ($\text{lr} = 10^{-3}$) and one for the classifier ($\text{lr} = 10^{-3}$, weight decay $= 10^{-4}$). Gradients were clipped at a global norm of 1.0. Each optimizer was paired with a ReduceLROnPlateau scheduler (patience $=20$ epochs).

\subsection*{Batching} 
Input data were provided as \texttt{AnnData} objects, which allow joint storage of feature matrices and sample-level metadata. The primary feature matrix (\texttt{.X}) was supplied either as a dense \texttt{NumPy} array or as a sparse matrix; sparse inputs were densified prior to conversion into \texttt{PyTorch} tensors and cast to \texttt{float32}. All batches were automatically transferred to the active device (CPU or GPU).  

Sample metadata were taken from the \texttt{.obs} field. To ensure compatibility with PyTorch, categorical columns were converted to numeric codes, and any multi-dimensional fields (e.g., one-hot encodings stored as matrices) were split into multiple one-dimensional columns with suffixes (e.g., \texttt{type\_0}, \texttt{type\_1}). Binary response labels were extracted from the \texttt{type} field; if the field contained two columns, the second column (\texttt{type\_1}) was interpreted as the positive class. This preprocessing ensured that each mini-batch contained both the feature tensor and a clean, aligned vector of binary response labels.

\subsection*{Logging} 
During training, we logged epoch-level training loss and ROC-AUC, as well as validation loss and ROC-AUC. Every 10 epochs, latent representations ($z$) were exported to TensorBoard for embedding visualization alongside associated sample metadata.

\subsection{Post-Training Analysis}
\subsection*{Feature Importance and Interpretation}
To interpret latent features, we used SHAP (SHapley Additive exPlanations) values computed on LightGBM classifiers trained on BDVAE embeddings. Latents with FDR $<0.05$ were retained. Pathway-level enrichment of significant latents was evaluated with permutation testing, and effect sizes were computed using standardized differences ($\delta$).  

\subsection*{Statistical Analyses}
Predictive performance was summarized using area under the ROC curve (AUC) and area under the precision-recall curve (AUPRC). Confidence intervals were estimated by bootstrapping (1,000 iterations). Kaplan–Meier survival curves were generated using the lifelines R package, with group differences tested by log-rank test. Multiple hypothesis testing was corrected by Benjamini–Hochberg FDR.  

\subsection{Software, Pipelines, and Versions}

\begin{table}[ht]
\centering
\caption{Terra Workflows and WDL Resources Used}
\renewcommand{\arraystretch}{1.3}
\begin{tabular}{|p{4cm}|p{7cm}|} 
\hline
\textbf{Workflow Name} & \textbf{Terra URL / WDL Resource} \\
\hline
Paired FASTQ to uBAM & \href{https://dockstore.org/workflows/github.com/gatk-workflows/seq-format-conversion/Paired-FASTQ-to-Unmapped-BAM:3.0.0?tab=info}{Link to Paired FASTQ to uBAM Workflow} \\
\hline
Preprocessing for Variant Discovery (HG38) & \href{https://dockstore.org/workflows/github.com/gatk-workflows/gatk4-data-processing/processing-for-variant-discovery-gatk4:2.1.0?tab=info}{Link to HG38 Mapping Workflow} \\
\hline
Mutect2 Panel of Normals (PoN) & \href{https://dockstore.org/workflows/github.com/broadinstitute/gatk/mutect2_pon:4.1.7.0?tab=info}{Link to Mutect2 PON Workflow} \\
\hline
Somatic Variant Calling with Mutect2 & \href{https://dockstore.org/workflows/github.com/broadinstitute/gatk/mutect2:4.1.7.0?tab=info}{Link to Somatic Variant Calling Workflow} \\
\hline
Variant Annotation (Funcotator + HGVS + OpenCravat) & \href{https://dockstore.org/workflows/github.com/IfrahTariq/analysis-wdls/detect_variants:main?tab=info}{Link to Variant Annotation Workflow} \\
\hline
RNA-seq Processing & \href{https://dockstore.org/workflows/github.com/IfrahTariq/bulkrnaseq/bulkRNA_workflow:main?tab=info}{Link to RNA-seq Processing Workflow} \\
\hline
\end{tabular}
\label{tab:terra_wdl_links}
\end{table}

\begin{table}[ht]
\centering
\caption{Core Software Libraries for Modeling and Analysis}
\begin{tabular}{|l|l|p{7cm}|}
\hline
\textbf{Library} & \textbf{Version} & \textbf{Application} \\
\hline
Python & 3.9 & Core implementation \\
PyTorch Lightning & 2.0+ & BDVAE model training \\
Optuna & 3.1 & Hyperparameter optimization \\
LightGBM & 3.3 & Downstream classification \\
scikit-learn & 1.2 & Model evaluation, preprocessing \\
scanpy & 1.9 & RNA-seq feature handling \\
lifelines & 0.27 & Survival analysis \\
GSEApy & 1.0 & Pathway curation \\
\hline
\end{tabular}
\label{tab:ml_versions}
\end{table}

\section*{Acknowledgements}
This work was supported by the Stand Up To Cancer (SU2C) Convergence 2.0 Grant and a fellowship to Ifrah Tariq from the Eric and Wendy Schmidt Center at the Broad Institute of MIT and Harvard. We thank colleagues in the Friedman, Wherry, and Barzilay labs, as well as the BioMicro Center, for helpful discussions and technical assistance. Computational resources were performed using the Terra platform (Broad Institute and Verily) and the MIT Engaging Cluster at the Massachusetts Green High Performance Computing Center (MGHPCC). 

\clearpage
\bibliographystyle{plain} 
\bibliography{biblios/fullbib}

\begin{thebibliography}{10}

\bibitem{Adinolfi2015}
Elena Adinolfi, Monica Capece, Anna Franceschini, Francesco Amoroso, Simonetta Falzoni, Anna~L Giuliani, Enrico Rotondo, Andrea~C Sarti, Massimo Bonora, and Francesco Di~Virgilio.
\newblock P2x7: a new player in inflammation and cancer.
\newblock {\em Oncogene}, 34(2):186--200, 2015.

\bibitem{Bathgate2013}
Ross~A.D. Bathgate, Michelle~L. Halls, Emma~T. van~der Westhuizen, Grant~E. Callander, Monika Kocan, and Roger~J. Summers.
\newblock Relaxin: new pathophysiological roles and clinical relevance.
\newblock {\em Trends in Pharmacological Sciences}, 34(1):11–20, 2013.

\bibitem{Bhat2009}
Ramesh Bhat and Lawrence Steinman.
\newblock Innate and adaptive autoimmunity directed to the central nervous system.
\newblock {\em Neuron}, 64(1):123--132, 2009.

\bibitem{Chang2015}
Chih-Hao Chang et~al.
\newblock Metabolic competition in the tumor microenvironment is a driver of cancer progression.
\newblock {\em Cell}, 162(6):1229--1241, 2015.

\bibitem{darvin2018immune}
Pradeep Darvin, Salman~M Toor, Varun Sasidharan~Nair, and Eyad Elkord.
\newblock Immune checkpoint inhibitors: recent progress and potential biomarkers.
\newblock {\em Experimental \& Molecular Medicine}, 50(12):1--11, 2018.

\bibitem{Dionisio2011}
Leonardo Dionisio, Maria Jose De~Rosa, Cecilia Bouzat, and Maria~C Esandi.
\newblock Gaba receptors and the immune system: New insights into neuroimmune interaction.
\newblock {\em CNS \& Neurological Disorders-Drug Targets}, 10(1):92--102, 2011.

\bibitem{elmarakeby2021pnet}
Haitham~A. Elmarakeby et~al.
\newblock Biologically informed deep neural network for prostate cancer discovery.
\newblock {\em Nature}, 598:348--352, 2021.

\bibitem{Fu2020}
Jingjing Fu, Kaixuan Li, Wenjie Zhang, Chunlong Wan, Jie Zhang, Peng Jiang, and X.~Shirley Liu.
\newblock Large-scale public data reuse to model immunotherapy response and resistance.
\newblock {\em Genome Medicine}, 12(1):21, Feb 2020.

\bibitem{greene2021interpretable}
Evan Greene et~al.
\newblock New interpretable machine-learning method for single-cell data reveals correlates of clinical response to cancer immunotherapy.
\newblock {\em Patterns (N Y)}, 2(12), 2021.

\bibitem{Ho2015}
Ping-Chih Ho, Justin~D Bihuniak, Andrew~N Macintyre, Matthew Staron, Xiao Liu, Raul Amezquita, Yvonne-Collin Tsui, Yufang Cui, Gregor Micevic, Jose~C Perales, et~al.
\newblock Phosphoenolpyruvate is a metabolic checkpoint of anti-tumor t cell responses.
\newblock {\em Cell}, 162(6):1217--1228, 2015.

\bibitem{hossain2024immune}
Md~Saiful~Islam Hossain et~al.
\newblock Immune checkpoint inhibitors in cancer therapy: mechanisms, biomarkers, and strategies for enhancing response.
\newblock {\em Cancer Letters}, 600:219--232, 2024.

\bibitem{hugo2016genomic}
Willy Hugo, Jesse~M Zaretsky, Lu~Sun, Chao Song, Brian~H Moreno, Siwen Hu-Lieskovan, Beata Berent-Maoz, Jae~Y Pang, Bartosz Chmielowski, George Cherry, et~al.
\newblock Genomic and transcriptomic features of response to anti-pd-1 therapy in metastatic melanoma.
\newblock {\em Science}, 352(6282):207--212, 2016.

\bibitem{Jin2013}
Zhe Jin, Sekhar~K Mendu, and Bryndis Birnir.
\newblock Gabaergic signaling in immune cells.
\newblock {\em Frontiers in Cellular Neuroscience}, 7:61, 2013.

\bibitem{Katagiri2002}
Kazuya Katagiri, Makoto Hattori, Nagahiro Minato, and Tatsuo Kinashi.
\newblock Rap1 is a key regulator of t-cell and antigen-presenting cell interactions and modulates t-cell responses.
\newblock {\em Molecular and Cellular Biology}, 22(3):1001–1015, 2002.

\bibitem{kim2018pembrolizumab}
Hyungsoon Kim, Yung-Jue Bang, Yoon-Koo Kang, Charles~S Fuchs, Shukui Qin, Takaki Satoh, Kohei Shitara, Josep Tabernero, Eric Van~Cutsem, Daniel~Y Oh, et~al.
\newblock Safety and efficacy of pembrolizumab in previously treated advanced gastric and gastroesophageal junction cancer: phase ii clinical keynote-059 trial.
\newblock {\em Nature Medicine}, 24(5):631--640, 2018.

\bibitem{kong2022netbio}
J.-H. Kong et~al.
\newblock A network-based machine learning framework identifies immunotherapy biomarkers and predicts immune checkpoint inhibitor response across multiple cancer types.
\newblock {\em Nature Communications}, 2022.

\bibitem{Leone2019}
Robert~D Leone, Lei Zhao, Jennifer~M Englert, Ivan~M Sun, Minsoo-Hong Oh, Irene~H Sun, Matthew~L Arwood, Isaac~A Bettencourt, Chetan~H Patel, Jing Wen, et~al.
\newblock Glutamine blockade induces divergent metabolic programs to overcome tumor immune evasion.
\newblock {\em Science}, 366(6468):1013--1021, 2019.

\bibitem{Li2022}
Xin Li, Wei Gu, Yu~Sun, and Wencan Yu.
\newblock Glp-1 receptor agonists modulate tumor immunity.
\newblock {\em Frontiers in Immunology}, 13:824276, 2022.

\bibitem{li2024review}
Y.~Li et~al.
\newblock Informing immunotherapy with multi-omics driven machine learning.
\newblock {\em NPJ Digital Medicine}, 2024.

\bibitem{Lieberman2003}
Judy Lieberman.
\newblock Granzyme a: a weapon of cytotoxic lymphocytes.
\newblock {\em Trends in Immunology}, 24(5):276--280, 2003.

\bibitem{mariathasan2018tgf}
Sanjeev Mariathasan, Shannon~J Turley, Derek Nickles, Alessandro Castiglioni, Kin-Ming Yuen, Yi~Wang, Elizabeth~E Kadel, Heidi Koeppen, Jillian~L Astarita, Rafael Cubas, et~al.
\newblock Tgf$\beta$ attenuates tumour response to pd-l1 blockade by contributing to exclusion of t cells.
\newblock {\em Nature}, 554(7693):544--548, 2018.

\bibitem{mcdermott2019atezolizumab}
David~F McDermott, Mohammed~A Huseni, Michael~B Atkins, Robert~J Motzer, Brian~I Rini, Bernard Escudier, Lawrence Fong, Ronald~W Joseph, Sumanta~K Pal, James~A Reeves, et~al.
\newblock Clinical activity and molecular correlates of response to atezolizumab alone or in combination with bevacizumab versus sunitinib in renal cell carcinoma.
\newblock {\em Nature Medicine}, 25(6):861--867, 2019.

\bibitem{Moore2003}
DJ~Moore, JK~Chambers, HJ~Waldvogel, RL~Faull, and PC~Emson.
\newblock P2y6 receptors mediate monocyte inflammatory responses to udp and lipopolysaccharide.
\newblock {\em Journal of Biological Chemistry}, 278(49):48580--48587, 2003.

\bibitem{riaz2017melanoma}
Na~Riaz, John~J Havel, Vladimir Makarov, Aur{\'e}lie Desrichard, Walter~J Urba, John~S Sims, F~Stephen Hodi, Pablo Mart{\'\i}nez, Alejandro Luna, Phil Wong, et~al.
\newblock Tumor and microenvironment evolution during immunotherapy with nivolumab.
\newblock {\em Cell}, 171(4):934--949.e16, 2017.

\bibitem{Scharping2021}
Nicholas~E Scharping, David~B Rivadeneira, Amanda~V Menk, Paolo D~A Vignali, Brandon~R Ford, Nicole~L Rittenhouse, Roberto~M Peralta, Yan Wang, Yao Wang, Kelsey DePeaux, et~al.
\newblock Mitochondrial stress induced by continuous stimulation underlies t cell dysfunction in tumors.
\newblock {\em Cell Reports}, 34(2):108590, 2021.

\bibitem{shen2025compass}
W.~Shen et~al.
\newblock Compass: a concept-bottleneck foundation model for interpretable pan-cancer immunotherapy response prediction.
\newblock {\em bioRxiv / medRxiv preprint}.

\bibitem{Takata2021}
Kaoru Takata, Yoshihiro Kitamura, Jun-ichiro Kakimura, Shin-ichiro Honda, and Shinya Suzu.
\newblock Glutamate signaling in immune cells.
\newblock {\em FEBS Journal}, 288(10):3348--3362, 2021.

\bibitem{vanallen2015genomic}
Eliezer~M Van~Allen, Diana Miao, Bastian Schilling, Sachet~A Shukla, Christian Blank, Lukas Zimmer, Antje Sucker, Ulrike Hillen, Marjolein~HG Foppen, Simone~M Goldinger, et~al.
\newblock Genomic correlates of response to ctla-4 blockade in metastatic melanoma.
\newblock {\em Science}, 350(6257):207--211, 2015.

\bibitem{vanguri2022multimodal}
R.~S. Vanguri et~al.
\newblock Multimodal integration of radiology, pathology and genomics enhances immunotherapy response prediction in nsclc.
\newblock {\em Nature Cancer}, 2022.

\bibitem{Wheeler2011}
David~W Wheeler, Antony~J Thompson, Federica Corletto, Janet Reckless, Yoon~K Loke, Nathalie Lapaque, Adel Boueiz, John Trowsdale, Geoffrey Bellingan, and Mervyn Singer.
\newblock Gaba-a receptors regulate innate immune functions.
\newblock {\em Journal of Neuroinflammation}, 8(1):108, 2011.

\end{thebibliography}

\clearpage
\appendix
\section*{Supplementary Figures}

\renewcommand{\thefigure}{S\arabic{figure}}
\renewcommand{\thetable}{S\arabic{table}}
\setcounter{figure}{0}
\setcounter{table}{0}

\begin{figure}[ht]
\centering
\includegraphics[width=0.95\textwidth]{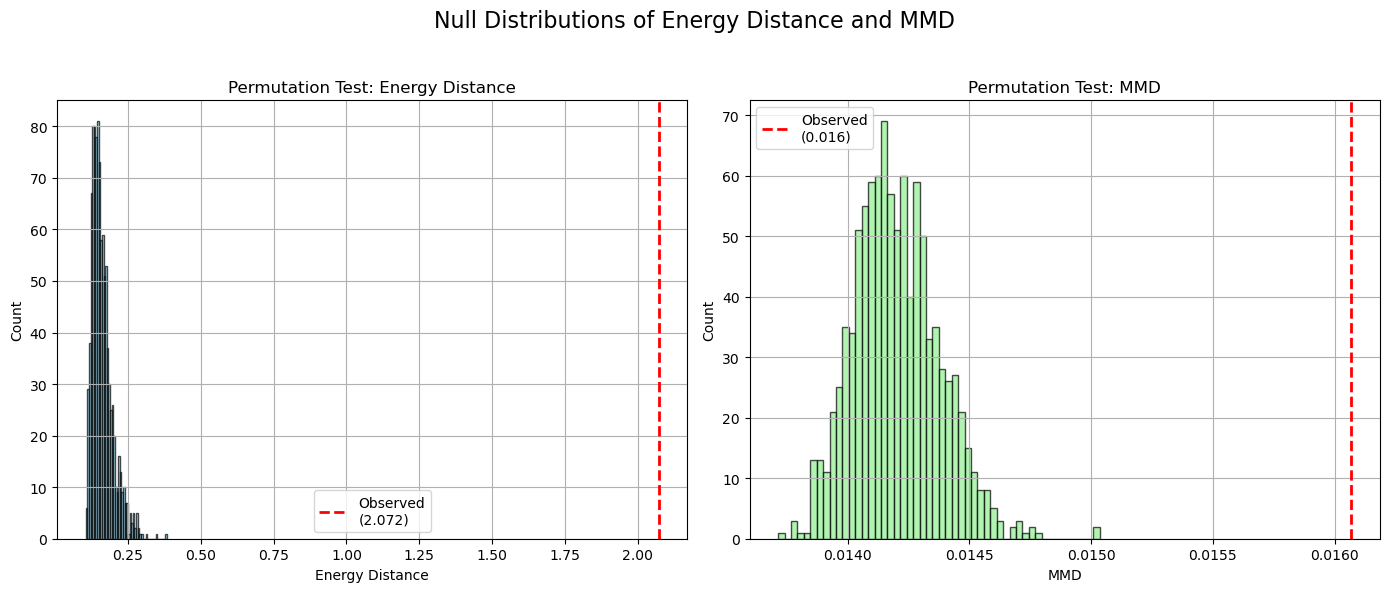}
\caption{
\textbf{Null distributions of Energy Distance and Maximum Mean Discrepancy (MMD).} 
Observed statistics (red dashed lines) far exceed the null distributions generated by 1,000 random label permutations, confirming that responders and non-responders occupy distinct regions of the BDVAE latent space.
}
\label{fig:supp_latent_permutation}
\end{figure}

\begin{figure}[ht]
    \centering
    \includegraphics[width=0.95\textwidth]{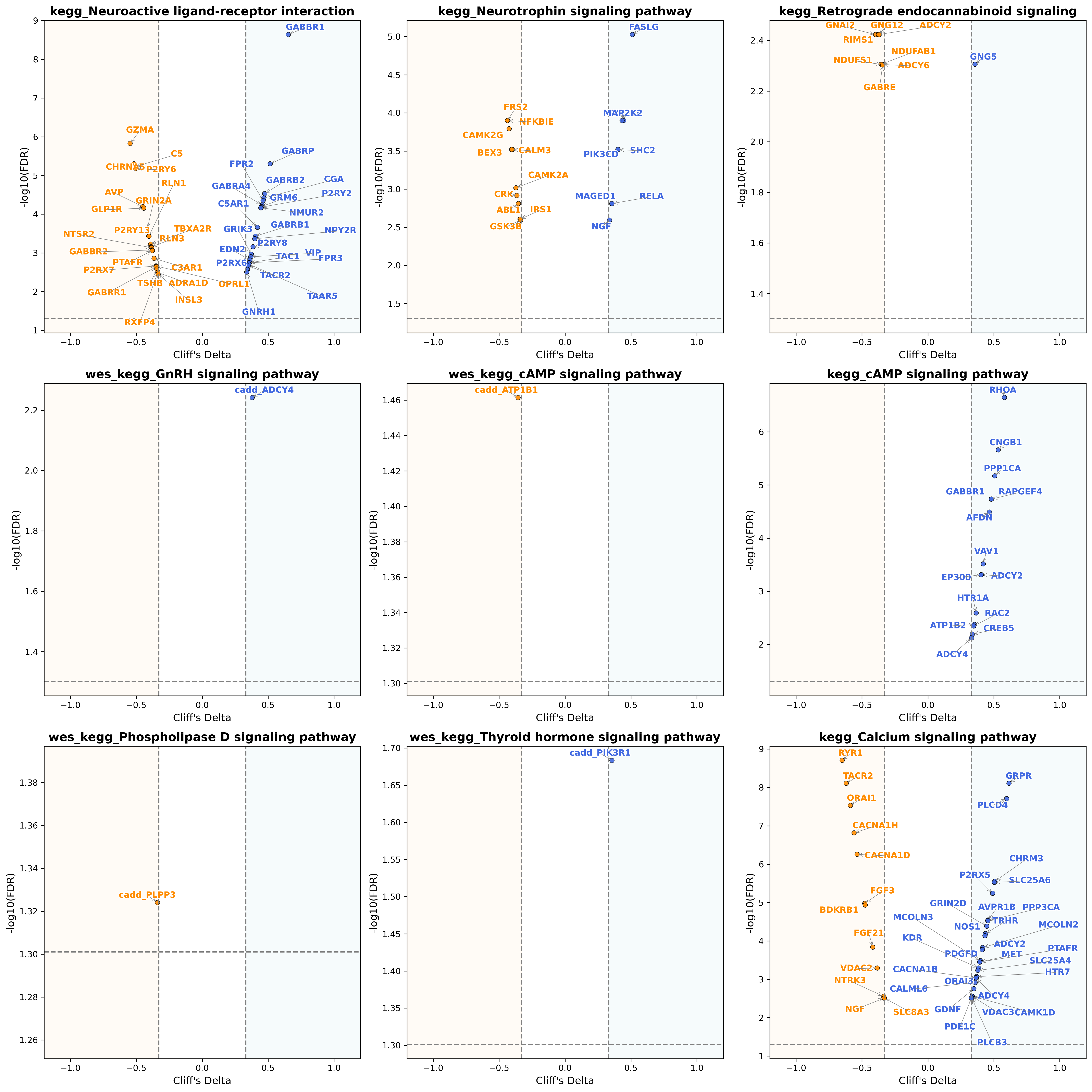}
    \caption{
    \textbf{Gene-level attribution differences in neuronal and neuroimmune pathways between responder- and non-responder–enriched clusters.} 
    Volcano plots summarize integrated gradient (IG) attributions for genes in selected pathways, comparing C1-NR (blue) and C2-R (orange). 
    The x-axis shows Cliff’s Delta effect size; the y-axis shows $-\log_{10}$(FDR). 
    Genes highlighted in orange contribute more strongly in responder-enriched tumors (C2-R), whereas those in blue are more influential in non-responder–enriched tumors (C1-NR).
    }
    \label{fig:neuro_pathway_volcano_grid}
\end{figure}

\begin{figure}[ht]
    \centering
    \includegraphics[width=0.95\textwidth]{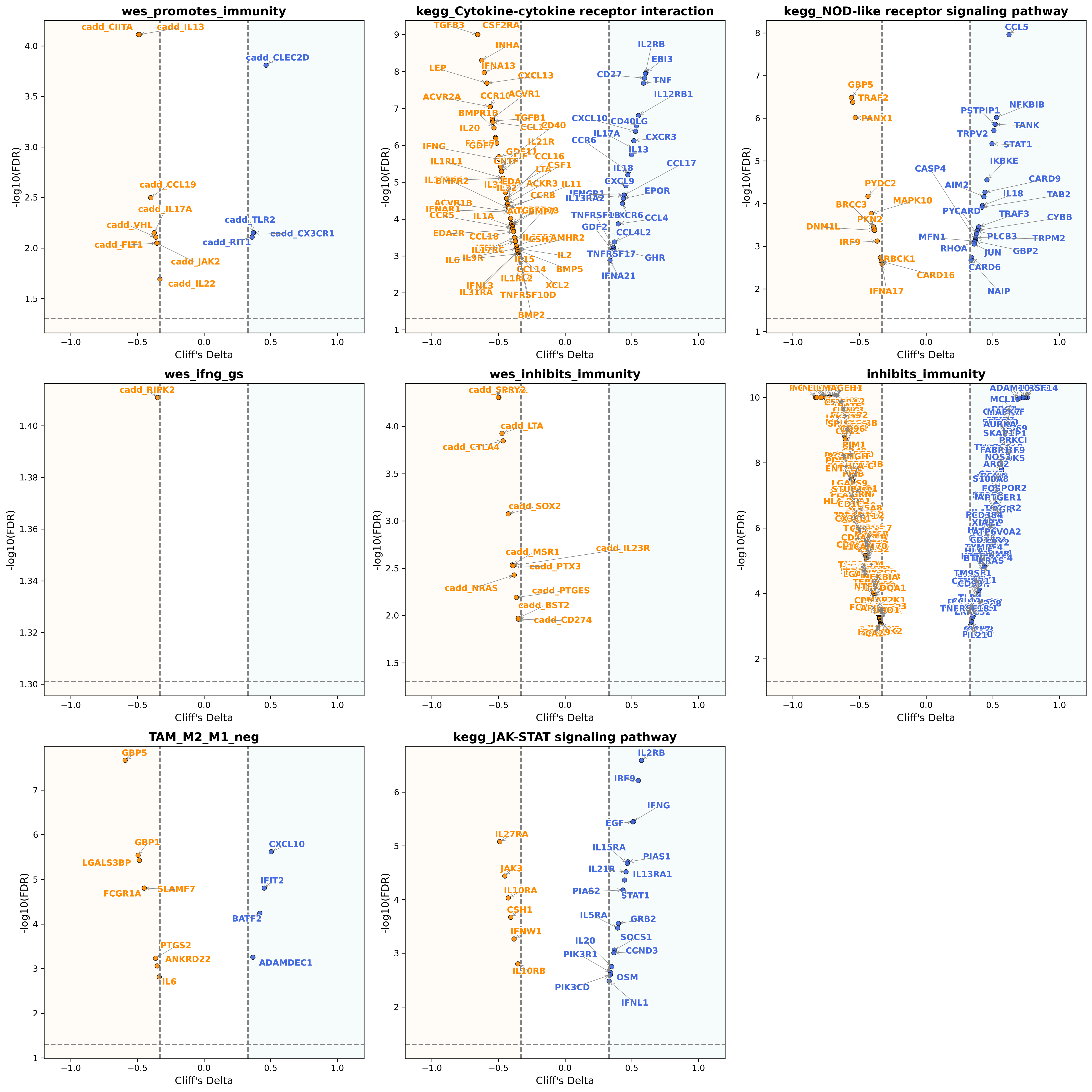}
    \caption{
    Gene-level gradient comparisons across selected neuronal and neuroimmune signaling pathways between C1-NR (non-responder dominant; blue) and C2-R (responder dominant; orange) clusters.
    Each subplot shows a volcano plot with Cliff’s Delta (x-axis) and $-\log_{10}$(FDR) (y-axis).
    Significant genes with integrated gradients enriched in responders are shown in orange; genes enriched in non-responders are shown in blue.
    }
    \label{fig:immune_volcano_panels}
\end{figure}

\begin{figure}[ht]
    \centering
    \includegraphics[width=\textwidth,height=0.8\textheight,keepaspectratio]{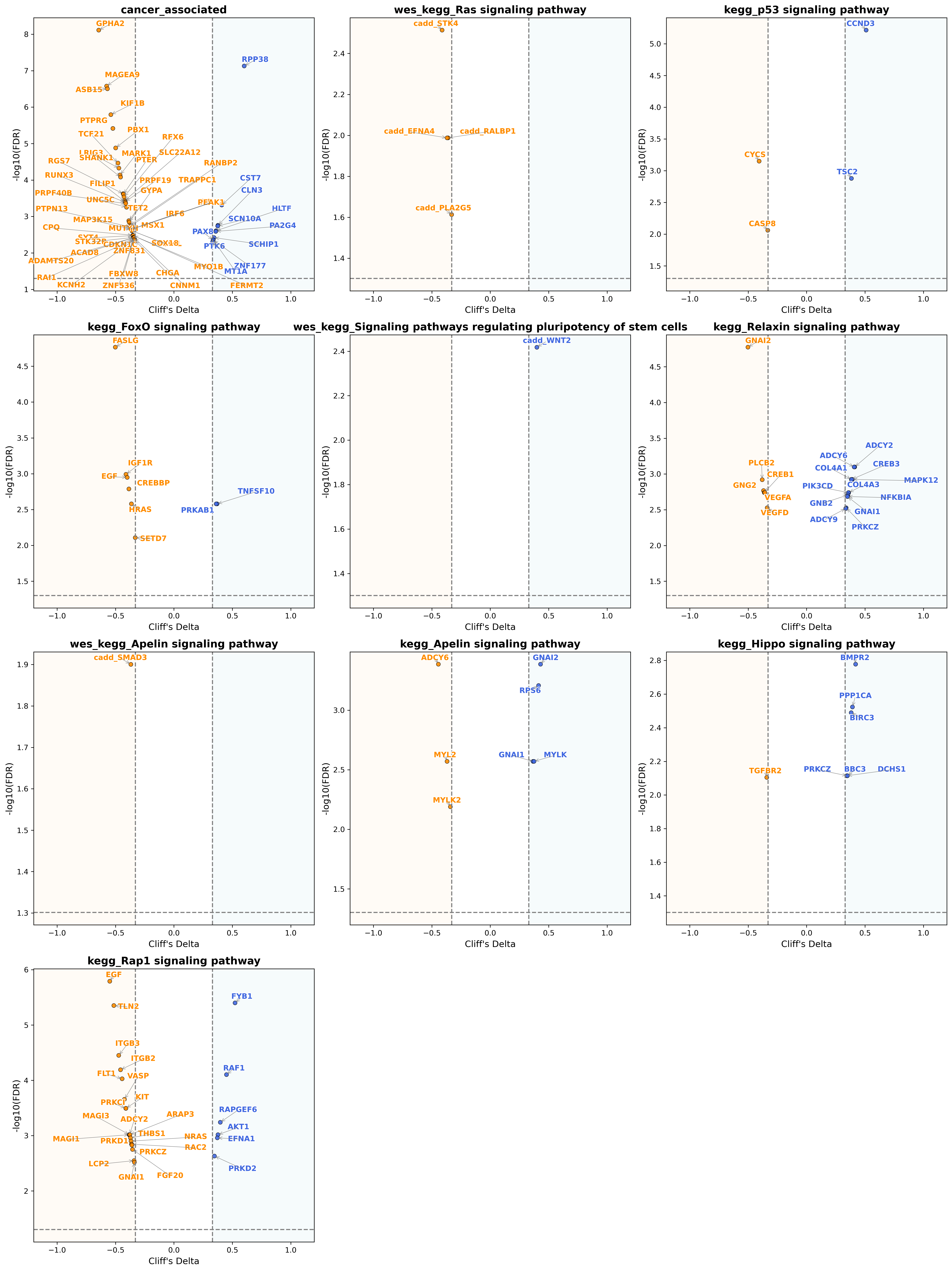}
    \caption{
    Volcano plots showing gene-level integrated gradients attribution across cancer-associated signaling pathways.
    Cliff’s Delta indicates the direction and magnitude of attribution differences between clinical groups, while $-\log_{10}$(FDR) reflects statistical confidence.
    Orange genes are more attributed to responder-enriched Cluster~2 (C2-R), while blue genes are more attributed to non-responder–dominant Cluster~1 (C1-NR).
    Pathways include Ras, p53, FoxO, Hippo, Relaxin, Rap1, stemness-related, and Apelin signaling.
    }
    \label{fig:cancer_volcano_panels}
\end{figure}

\begin{figure}[ht]
    \centering
    \includegraphics[width=\textwidth]{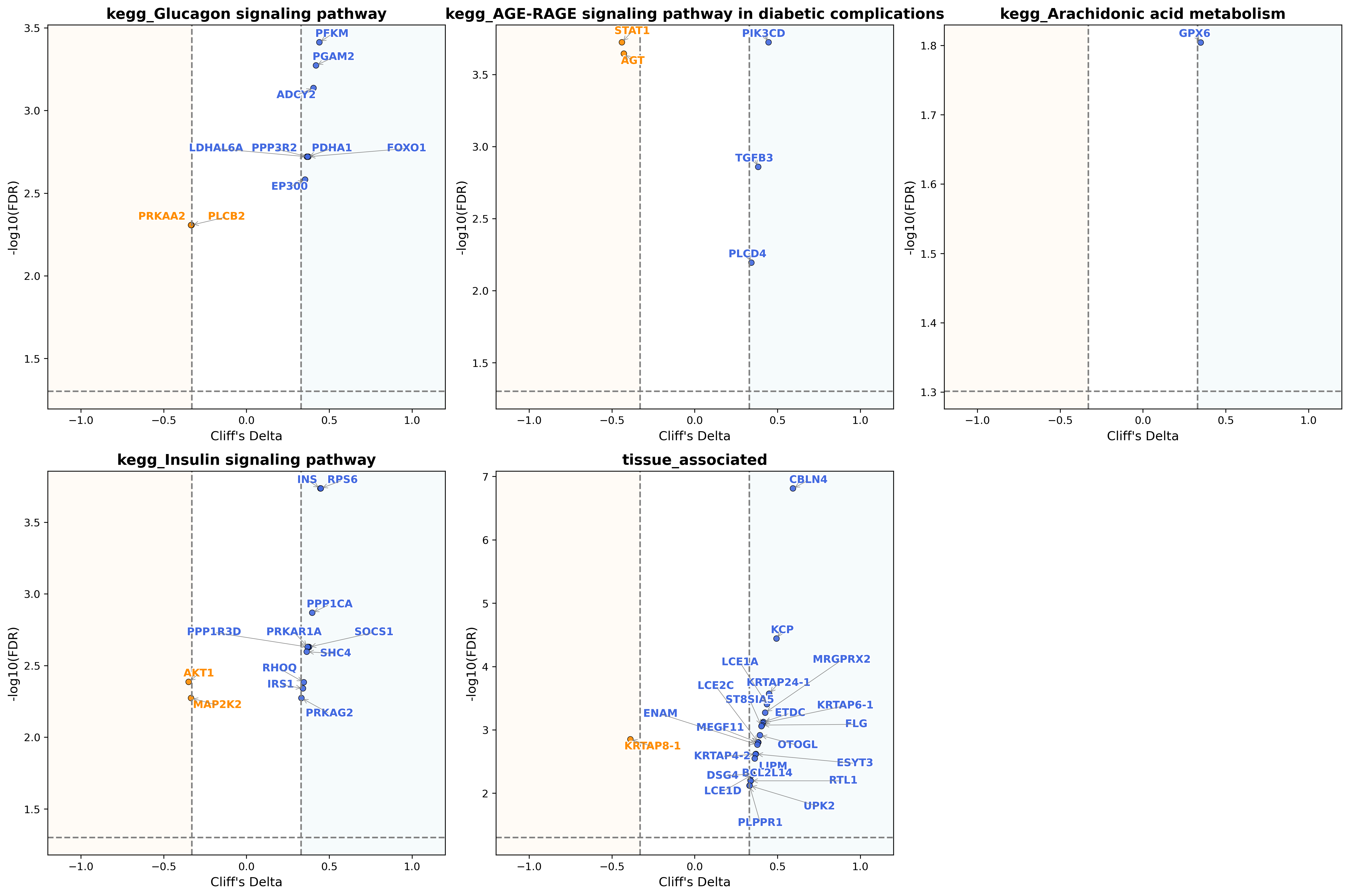}
    \caption{
    Volcano plots showing integrated gradients attribution for metabolic and tissue-associated pathways.
    Genes are colored by directional enrichment toward responder (orange) or non-responder (blue) dominant clusters.
    Pathways include \textit{Glucagon}, \textit{Insulin}, and \textit{Arachidonic acid metabolism}, as well as a tissue-associated signature.
    Significant differences in attribution suggest cluster-specific metabolic programming and tissue remodeling states.
    }
    \label{fig:metabolism_volcano_panels}
\end{figure}

\begin{table}[htbp]
\centering
\caption{Immune signature encoders (TIDE-based)}
\renewcommand{\arraystretch}{1.2}
\begin{tabular}{|l|p{10cm}|}
\hline
\textbf{Encoder} & \textbf{Description} \\
\hline
\texttt{apm} & Antigen presentation machinery genes influencing MHC class I expression and immune recognition \\
\texttt{isg\_rs} & Interferon-stimulated gene response signature \\
\texttt{ifng\_gs} & Interferon gamma signaling gene set \\
\texttt{T\_accum\_pos/neg} & T cell accumulation signatures (positive or negative direction) \\
\texttt{T\_exhaust\_pos/neg} & T cell exhaustion markers indicating chronic stimulation and dysfunction \\
\texttt{T\_exhaust.fixed\_pos/neg} & Canonical exhaustion gene set \\
\texttt{T\_regulatory\_pos/neg} & Regulatory T cell (Treg) marker signatures \\
\texttt{ICB\_resist\_pos/neg} & Genes linked to immune checkpoint blockade resistance mechanisms \\
\texttt{MDSC\_pos/neg} & Myeloid-derived suppressor cell markers \\
\texttt{TAM\_M2\_M1\_pos/neg} & Tumor-associated macrophage polarization markers (M2 vs. M1) \\
\texttt{CAF\_pos/neg} & Cancer-associated fibroblast signatures \\
\hline
\end{tabular}
\label{tab:supp_tide_encoders}
\end{table}

\begin{table}[htbp]
\centering
\caption{Cancer-associated encoders}
\renewcommand{\arraystretch}{1.2}
\begin{tabular}{|l|p{8cm}|}
\hline
\textbf{Source} & \textbf{Description} \\
\hline
COSMIC (CGC)  & Genes from the Cancer Gene Census filtered by tumor-specific keywords (e.g., ``renal'', ``melanoma'') \\
DisGeNET & Genes associated with disease terms like ``bladder cancer'', ``renal cell carcinoma'' \\
IntOGen & Genes with somatic mutation enrichment in specific tumors (e.g., SKCM, BLCA) \\
MalaCards &  Genes mapped to diseases from literature-curated associations; filtered by inclusion and optional exclusion terms \\
\hline
\end{tabular}
\label{tab:supp_cancer_encoders}
\end{table}

\begin{table}[htbp]
\centering
\caption{WES-derived feature groups}
\renewcommand{\arraystretch}{1.2}
\begin{tabular}{|l|c|p{7.5cm}|}
\hline
\textbf{Feature Group}  & \textbf{Count} & \textbf{Description} \\
\hline
Drug response features  & 40 & Binary indicators for the presence of variants known to influence drug response, extracted via \texttt{maftools}. \\
SBS signature features & 67 & Mutation signature features based on single base substitutions (SBS), derived from mutational signature analysis using \texttt{maftools}. \\
Oncogenic summary stats  & 11 & Includes tumor mutational burden (TMB) and oncogenic mutation burden scores, such as those beginning with \texttt{oncogenic\_}, derived from \texttt{maftools}. \\
CADD-based annotations  & 18,255 & Features based on Combined Annotation Dependent Depletion (CADD) scores for assessing the deleteriousness of single nucleotide variants. \\
\hline
\end{tabular}
\label{tab:supp_wes_features}
\end{table}

\end{document}